\documentclass[letterpaper,twocolumn,10pt]{article}

\usepackage{usenix,epsfig,endnotes}
\usepackage{booktabs}
\usepackage{subcaption}
\usepackage{authblk}
\usepackage{graphicx}
\usepackage{amsmath,amssymb}
\usepackage{array}
\usepackage{listings}
\usepackage{color}
\usepackage{cite}
\usepackage{nowidow}
\usepackage{textcomp}
\usepackage{xspace}
\usepackage{enumitem}
\usepackage[ruled]{algorithm2e}
\usepackage{hyperref}
\usepackage{times}
\usepackage{lipsum}
\usepackage{fancyhdr}
\newcommand{\precap}{\vskip -0mm}
\newcommand{\postcap}{\vskip -0mm}
\newcommand{\presec}{}
\newcommand{\postsec}{}

\newtheorem{thm:thm}{Theorem}[section]
\newtheorem{thm:def}{Definition}[section]
\newtheorem{thm:lemma}{Lemma}[section]

\newcommand{\squishlist}{
   \begin{list}{$\bullet$}
    { \setlength{\itemsep}{1pt}      \setlength{\parsep}{3pt}
      \setlength{\topsep}{3pt}       \setlength{\partopsep}{0pt}
      \setlength{\leftmargin}{1em} \setlength{\labelwidth}{1em}
      \setlength{\labelsep}{0.5em} } }

\newcommand{\squishlisttwo}{
   \begin{list}{$\bullet$}
    { \setlength{\itemsep}{0pt}    \setlength{\parsep}{0pt}
      \setlength{\topsep}{0pt}     \setlength{\partopsep}{0pt}
      \setlength{\leftmargin}{2em} \setlength{\labelwidth}{1.5em}
      \setlength{\labelsep}{0.5em} } }

\newcommand{\squishend}{
    \end{list}  }

\newcommand{\REV}[1]{#1}

\newcommand{\capsize}[1]{#1}

\newcommand{\TensorOpt}{TVM\xspace}
\newcommand{\TensorOptno}{TVM}

\newcommand{\accel}{VDLA\xspace}
\newcommand{\accelno}{VDLA\xspace}

\newcommand{\accelmeaning}{Vanilla Deep Learning Accelerator}

\begin{document}

%don't want date printed
\date{}

%% Title information
\title{TVM: An Automated End-to-End Optimizing Compiler for Deep Learning}

%% Author with single affiliation.
\author{
  \small{
  Tianqi Chen$ ^1$, \ \  Thierry Moreau$ ^1$,\ \ Ziheng Jiang$ ^{1, 2}$, \ \  Lianmin Zheng$ ^{3}$, \ \ Eddie Yan$ ^1$\\
  Meghan Cowan$ ^1$, \ \ Haichen Shen$ ^1$, \ \ \  Leyuan Wang$ ^{4, 2}$, \ \ \  Yuwei Hu$ ^5$,
  Luis Ceze$ ^1$, \ \ \  Carlos Guestrin$ ^1$, \ \ \ Arvind Krishnamurthy$ ^1$}

  \small{$ ^1$Paul G. Allen School of Computer Science \& Engineering, University of Washington}\\
  \small{$ ^2$ AWS,\ $ ^3$Shanghai Jiao Tong University,\ $ ^4$UC Davis, \ $ ^5$Cornell}
}
\maketitle

% Use the following at camera-ready time to suppress page numbers.
% Comment it out when you first submit the paper for review.

\subsection*{Abstract}
There is an increasing need to bring machine learning to a wide diversity of hardware devices.
Current frameworks rely on vendor-specific operator libraries and optimize for a narrow range of server-class GPUs.  Deploying workloads to new platforms -- such as mobile phones, embedded devices, and accelerators (e.g., FPGAs, ASICs) -- requires significant manual effort.  We propose \TensorOptno, a compiler that exposes graph-level and operator-level optimizations to provide performance portability to deep learning workloads across diverse hardware back-ends.  \TensorOpt solves optimization challenges specific to deep learning, such as high-level operator fusion, mapping to arbitrary hardware primitives, and memory latency hiding. It also automates optimization of low-level programs to hardware characteristics by employing a novel, learning-based cost modeling method for rapid exploration of code optimizations.
Experimental results show that \TensorOpt delivers performance across hardware back-ends that are competitive with state-of-the-art, hand-tuned libraries for low-power CPU, mobile GPU, and server-class GPUs.
We also demonstrate \TensorOptno's ability to target new accelerator back-ends, such as the FPGA-based generic deep learning accelerator.
The system is open sourced and in production use inside several major companies.

\presec
\section{Introduction}
\begin{figure}[t]
	\centering
	\includegraphics[width=.9\columnwidth]{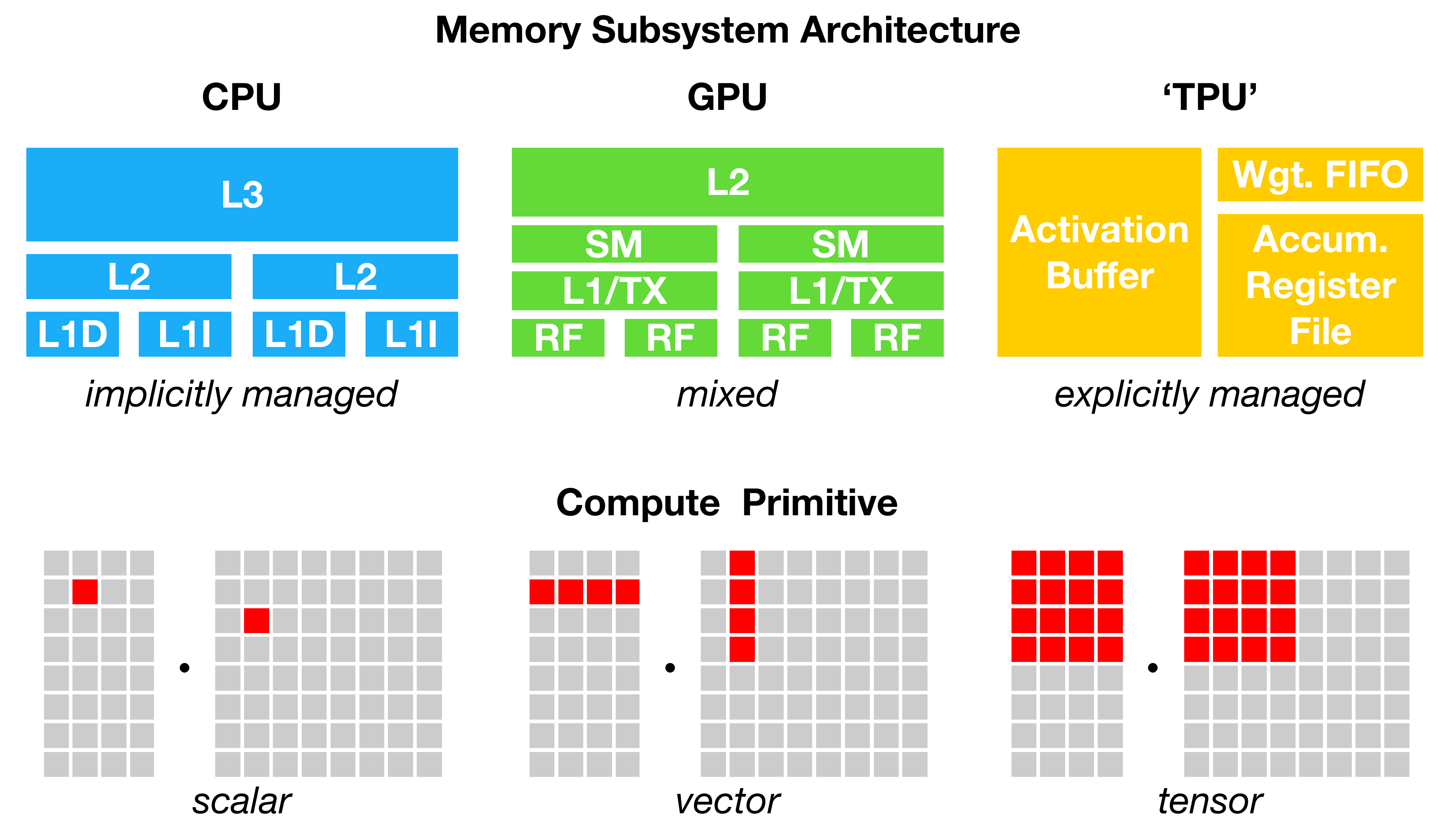}
	\precap
	\caption{CPU, GPU and TPU-like accelerators require different
		on-chip memory architectures and compute
		primitives. This divergence must be addressed when generating optimized code.}
	\postcap
	\label{fig:divergence}
\end{figure}

\postsec
\label{sec:intro}
Deep learning (DL) models can now recognize images, process natural language, and defeat humans in challenging strategy games. There is a growing demand to deploy smart applications to a wide spectrum of devices, ranging from cloud servers to self-driving cars and embedded devices. Mapping DL workloads to these devices is complicated by  the diversity of hardware characteristics, including embedded CPUs, GPUs, FPGAs, and ASICs (e.g., the TPU~\cite{Jouppi:TPU}). These hardware targets diverge in terms of memory organization, compute functional units, etc.,  as shown in \autoref{fig:divergence}.

Current DL frameworks, such as TensorFlow, MXNet, Caffe, and PyTorch, rely on a computational graph intermediate representation to implement optimizations, e.g., auto differentiation and dynamic memory management~\cite{tensorflow-osdi,MXNet-whitepaper,CNTK}. Graph-level optimizations, however, are often too high-level to handle hardware back-end-specific operator-level transformations. Most of these frameworks focus on a narrow class of server-class GPU devices and delegate target-specific optimizations to highly engineered and vendor-specific operator libraries. These operator-level libraries require significant manual tuning and hence are too specialized and opaque to be easily ported across hardware devices. Providing support in various DL frameworks for diverse hardware back-ends presently requires significant engineering effort.  Even for supported back-ends, frameworks must make the difficult choice between: (1) avoiding graph optimizations that yield new operators not in the predefined operator library, and (2) using unoptimized implementations of these new operators.

To enable both graph- and operator-level optimizations for diverse hardware back-ends, we take a fundamentally different, end-to-end
approach. \textit{We built \TensorOpt, a compiler that takes a high-level specification of a deep learning program from existing frameworks and generates low-level optimized code for a diverse set of hardware back-ends. }To be attractive to users, \TensorOpt needs to offer performance competitive with the multitude of manually optimized operator libraries across diverse hardware back-ends. This goal requires addressing the key challenges described below.

\vspace{0.5ex} \noindent {\bf Leveraging Specific Hardware Features and Abstractions.}
DL accelerators introduce optimized tensor compute primitives
~\cite{Jouppi:TPU,volta-whitepaper,Chen:Eyeriss}, while GPUs and CPUs continuously improve their processing elements. This poses a significant challenge in generating optimized code for a given operator description.
The inputs to hardware instructions are multi-dimensional, with fixed or variable lengths; they dictate different data layouts; and they have special requirements for memory hierarchy.  The system must effectively exploit these complex primitives to benefit from acceleration. Further, accelerator designs also commonly favor leaner control~\cite{Jouppi:TPU} and offload most scheduling complexity to the compiler stack. For specialized accelerators, the system now needs to generate code that explicitly controls pipeline dependencies to hide memory access latency -- a job that hardware performs for CPUs and GPUs.

\vspace{0.5ex} \noindent {\bf Large Search Space for Optimization}
Another challenge is producing efficient code without manually tuning operators.  The combinatorial choices of memory access, threading pattern, and novel hardware primitives creates a huge configuration space for generated code (e.g., loop tiles and ordering, caching, unrolling) that would incur a large search cost if we implement black box auto-tuning. One could adopt a predefined cost model to guide the search, but building an accurate cost model is difficult due to the increasing complexity of modern hardware. Furthermore, such an approach would require us to build separate cost models for each hardware type.

\TensorOpt addresses these challenges with three key modules.
(\textbf{1}) We introduce a \textit{tensor expression language} to build operators and provide program transformation primitives that generate  different versions of the program with various optimizations. This layer extends Halide~\cite{JRK:Halide}'s compute/schedule separation concept by also separating target hardware intrinsics from transformation primitives, which enables support for novel accelerators and their corresponding new intrinsics. Moreover, we introduce new transformation primitives to address GPU-related challenges and enable deployment to specialized accelerators. We can then apply different sequences of program transformations to form a rich space of valid programs for a given operator declaration. (\textbf{2}) We introduce an \textit{automated program optimization framework} to find optimized tensor operators. The optimizer is guided by an ML-based cost model that adapts and improves as we collect more data from a hardware back-end. (\textbf{3}) On top of the automatic code generator, we introduce a \textit{graph rewriter} that takes full advantage of high-  and operator-level optimizations.

By combining these three modules, \TensorOpt can take model descriptions from existing deep learning frameworks, perform joint high- and low-level optimizations, and generate hardware-specific optimized code for back-ends, e.g., CPUs, GPUs, and FPGA-based specialized accelerators.

This paper makes the following contributions:
%\begin{itemize}
\squishlist
\item We identify the major optimization challenges in providing performance portability to deep learning workloads across diverse hardware back-ends.
\item We introduce novel schedule primitives that take advantage of cross-thread memory reuse, novel hardware intrinsics, and latency hiding.
\item We propose and implement a machine learning based optimization system to automatically explore and search for optimized tensor operators.
\item We build an end-to-end compilation and optimization stack that allows the deployment of deep learning workloads specified in high-level frameworks (including TensorFlow, MXNet, PyTorch, Keras, CNTK) to diverse hardware back-ends (including CPUs, server GPUs, mobile GPUs, and FPGA-based accelerators). The open-sourced \TensorOpt is in production use inside several major companies.

We evaluated \TensorOpt using real world workloads on a server-class GPU, an embedded GPU, an embedded CPU, and a custom generic FPGA-based accelerator. Experimental results show that \TensorOpt offers portable performance across back-ends and achieves speedups ranging from 1.2$\times$ to 3.8$\times$ over existing frameworks backed by hand-optimized libraries.
%\end{itemize}
\squishend

%Our system generates deployable code that is performance-competitive with state-of-the-art vendor-specific libraries,
%and can target new specialized accelerator back-ends.
\section{Overview}
\begin{figure}[t]
	\centering
  \includegraphics[width=1\columnwidth]{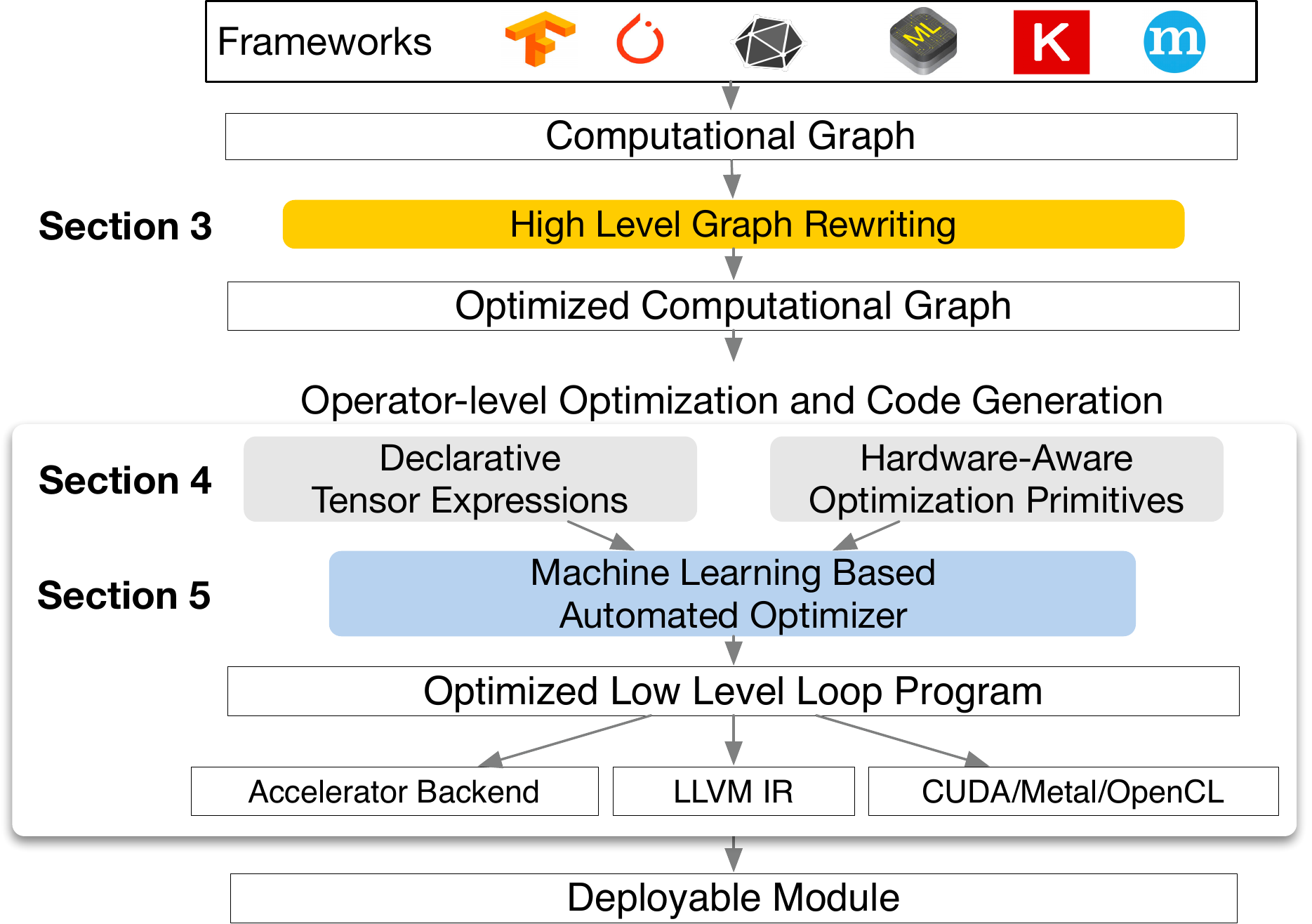}
	\precap
	\caption{System overview of \TensorOpt.
    The current stack supports descriptions from many deep learning
    frameworks and \REV{exchange formats, such as CoreML and ONNX,} to target major CPU, GPU and specialized accelerators.}
	\postcap
	\label{fig:overview}
\end{figure}

This section describes \TensorOpt by using an example to walk through
its components.
\autoref{fig:overview} summarizes execution steps in \TensorOpt and their corresponding sections in the paper.
The system first takes as input a model from an existing framework and transforms it into a computational graph representation.
It then performs high-level dataflow rewriting to generate an optimized graph.
The operator-level optimization module must generate efficient code for each fused operator in this graph.
Operators are specified in a declarative tensor expression language; execution details are unspecified.
\TensorOpt identifies a collection of possible code optimizations for a given hardware target's operators. Possible optimizations form a large space, so
we use an ML-based cost model to find optimized operators.
Finally, the system packs the generated code into a deployable module.

\paragraph{End-User Example.}
In a few lines of code, a user can take a model from existing deep learning frameworks
and call the \TensorOpt API to get a deployable module:

\includegraphics[scale=0.42]{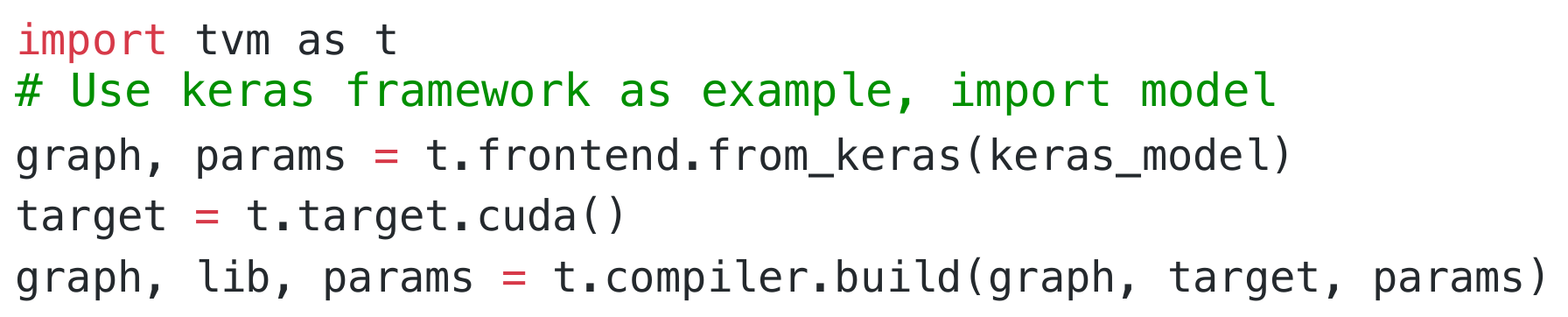}

This compiled runtime module contains three components:
the final optimized computational graph~(\texttt{graph}),
generated operators~(\texttt{lib}),
and module parameters~(\texttt{params}).
These components can then be used to deploy the model to the target back-end:

\includegraphics[scale=0.42]{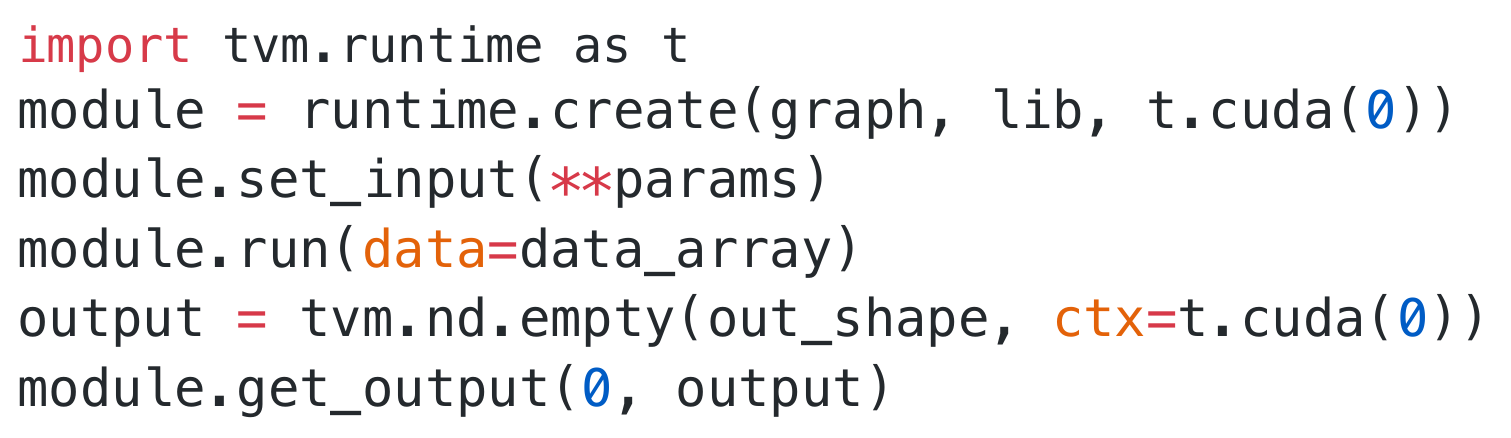}

\TensorOpt supports multiple deployment back-ends in languages such as C++, Java and
Python. The rest of this paper describes \TensorOpt's architecture and how a system programmer can extend it to support new back-ends.

\section{Optimizing Computational Graphs}
\label{sec:graph}

\begin{figure}[t]
\centering
\includegraphics[width=1\columnwidth]{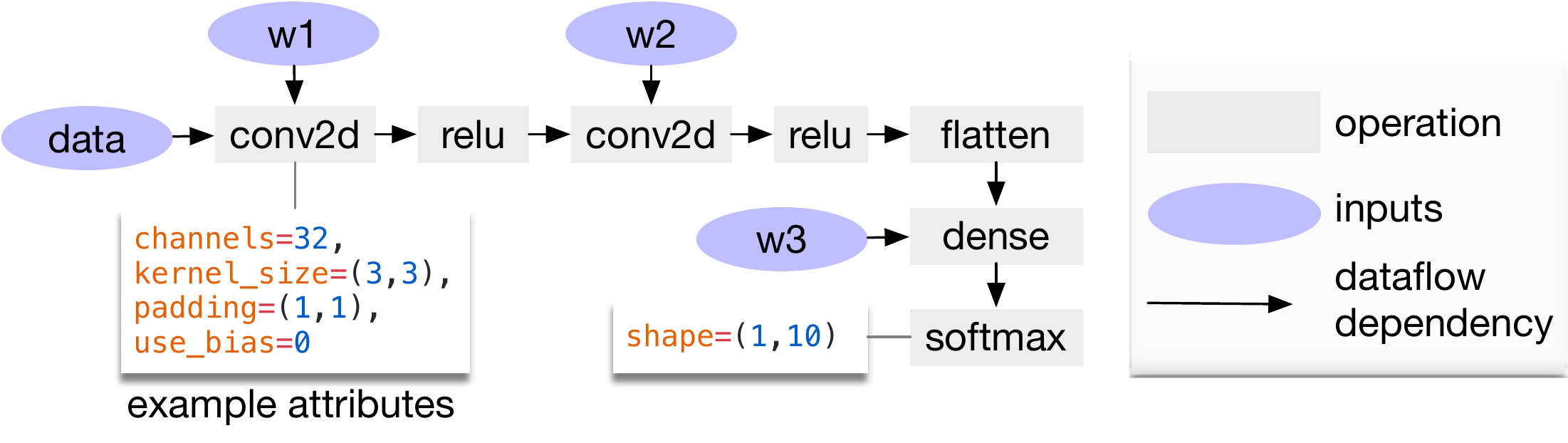}
\precap
\caption{Example computational graph of a two-layer convolutional neural network.
         Each node in the graph represents an operation that consumes one or more tensors
         and produces one or more tensors.
         Tensor operations can be parameterized by attributes to configure their behavior (e.g., padding or strides).}
\label{fig:dataflow}
\postcap
\end{figure}

Computational graphs are a common way to represent programs
in DL frameworks~\cite{tensorflow-osdi,Bastien-Theano-2012,MXNet-whitepaper,CNTK}.
~\autoref{fig:dataflow} shows an example computational graph
representation of a two-layer convolutional neural network.
The main difference between this high-level representation and a low-level compiler intermediate representation (IR), such as LLVM,
is that the intermediate data items are large, multi-dimensional tensors.
Computational graphs provide a global view of operators, but they avoid specifying how each operator must be implemented.
Like LLVM IRs, a computational graph can be transformed into functionally equivalent
graphs to apply optimizations.
\REV{We also take advantage of shape specificity in common DL workloads to optimize
for a fixed set of input shapes.}

\TensorOpt exploits a computational graph representation to apply high-level optimizations:
a node represents an operation on tensors or program inputs,
and edges represent data dependencies between operations.
It implements many graph-level optimizations, including: \textit{operator fusion}, which fuses multiple small operations together; \textit{constant-folding}, which pre-computes graph parts that can be determined statically, saving execution costs; a \textit{static memory planning pass}, which pre-allocates memory to hold each intermediate tensor; and \textit{data layout transformations}, which transform
%\CMT{is "massage" the right word?  Would "transform" be better?}
internal data layouts into back-end-friendly forms. We now discuss operator fusion and the data layout transformation.
%\CMT{Why do you pick only two of the four optimizations to discuss in further detail? tqchen: mainly because they are considered more novel contributions}

\begin{figure}[t]
 \centering
 \includegraphics[width=.95\columnwidth]{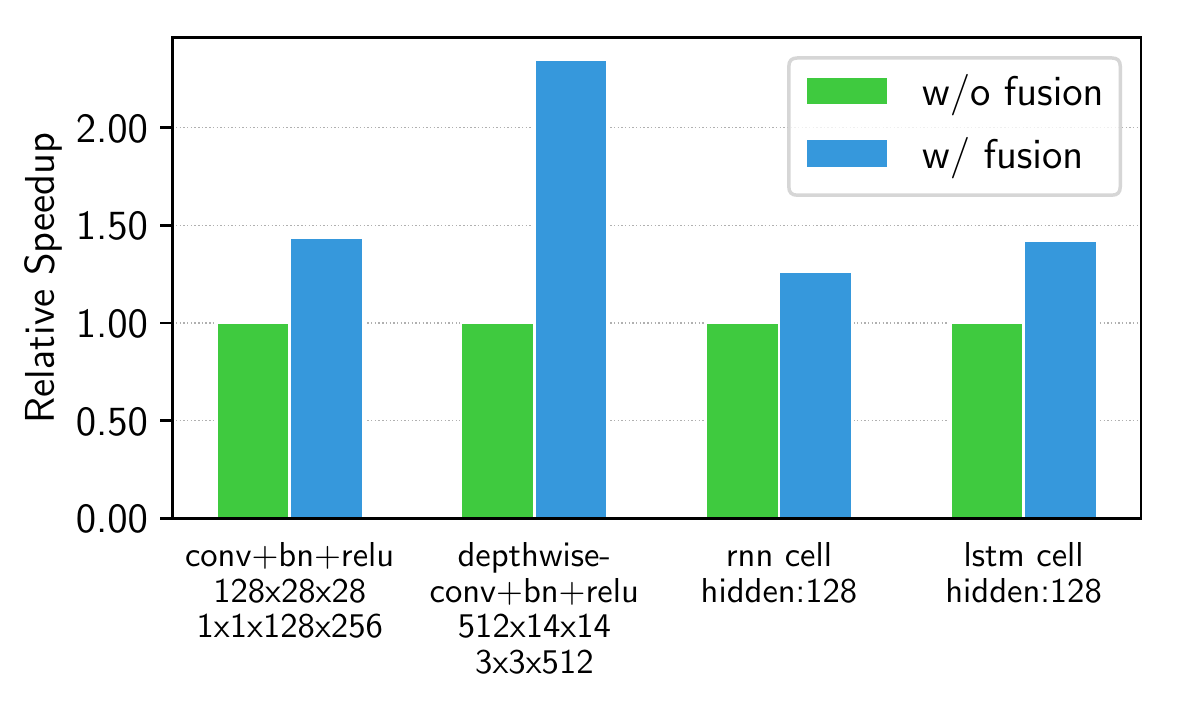}
 \precap
 \caption{Performance comparison between fused and non-fused operations. \TensorOpt generates both operations. \REV{Tested on NVIDIA Titan X.}
}
 \postcap
 \label{fig:fusion-cmp}
\end{figure}
%\begin{figure}[!t]
% \centering
% \includegraphics[width=1\columnwidth]{data_layout}
% \precap
% \caption{Data layout requirements can be affected by tensorization: here a 2$\times$2 tensorized operation dictates a data layout transformation.}
% \postcap
% \label{fig:datalayout}
%\end{figure}

\paragraph{Operator Fusion.}
Operator fusion combines multiple operators into a single kernel without saving the intermediate results in memory.
This optimization can greatly reduce execution time, particularly in GPUs and specialized accelerators.
Specifically, we recognize four categories of graph operators: (1) injective (one-to-one map, e.g., add), (2) reduction (e.g., sum), (3) complex-out-fusable (can fuse element-wise map to output, e.g., conv2d), and (4) opaque (cannot be fused, e.g., sort).
We provide generic rules to fuse these operators, as follows. Multiple injective operators can be fused into another injective operator. A reduction operator can be fused with input injective operators (e.g., fuse scale and sum). Operators such as conv2d are complex-out-fusable, and we can fuse element-wise operators to its output. We can apply these rules to transform the computational graph into a fused version.  ~\autoref{fig:fusion-cmp} demonstrates the impact of this optimization on different workloads.
We find that fused operators generate up to a 1.2$\times$ to 2$\times$ speedup by reducing memory accesses.

\paragraph{Data Layout Transformation.}
There are multiple ways to store a given tensor in the computational graph. The most common data layout choices are column major and row major. In practice, we may prefer to use even more complicated data layouts. For instance, a DL accelerator might exploit $4 \times 4$ matrix operations, requiring data to be tiled into $4 \times 4$ chunks to optimize for access locality.

Data layout optimization converts a computational graph into one that can use better internal data layouts for execution on the target hardware. It starts by specifying the preferred data layout for each operator given the constraints dictated by memory hierarchies. We then perform the proper layout transformation between a producer and a consumer if their preferred data layouts do not match.

\begin{figure}[t]
\centering
\includegraphics[width=.99\columnwidth]{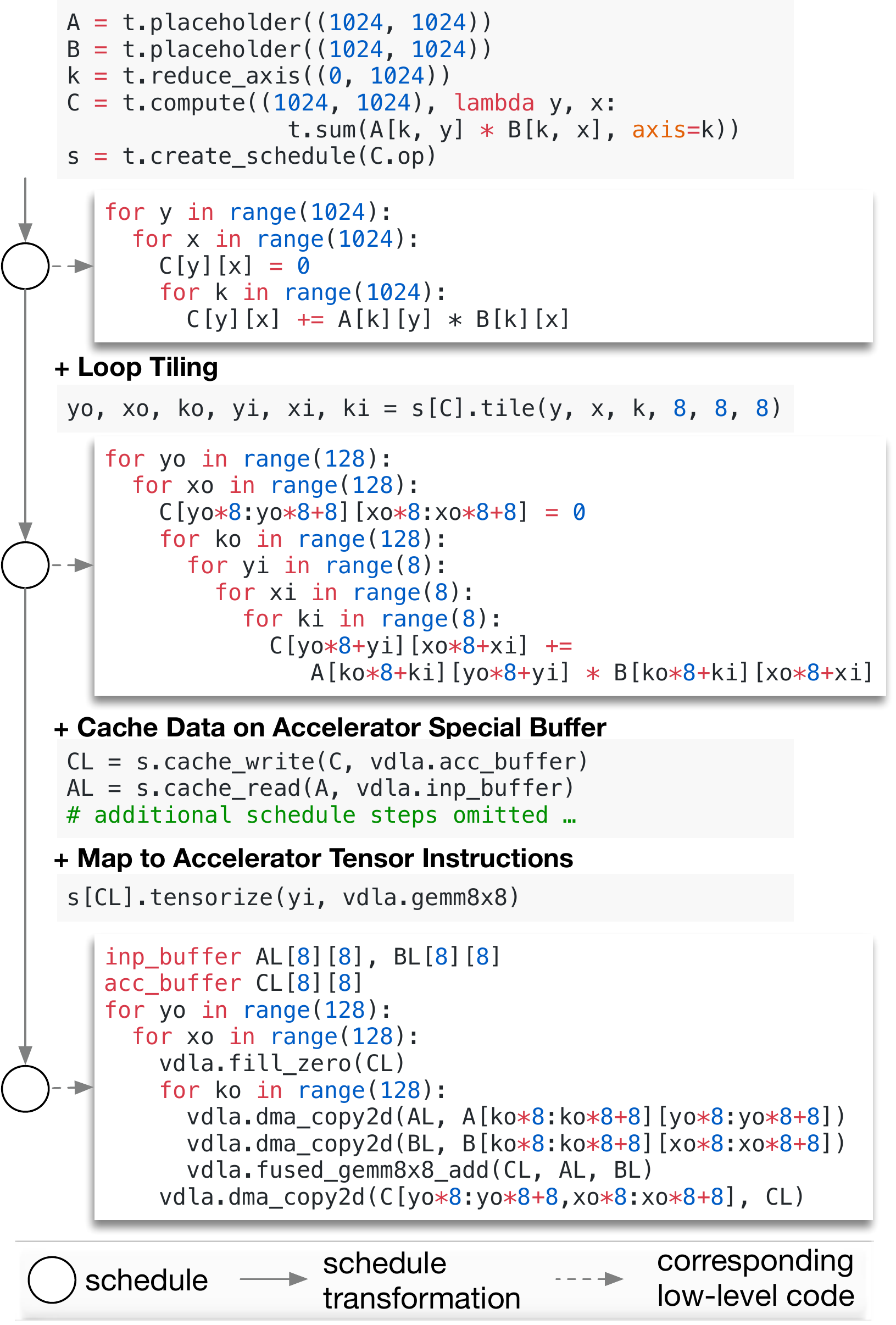}
\precap
\caption{Example schedule transformations that optimize a matrix multiplication on a specialized accelerator.}
\label{fig:gemm-flow}
\postcap
\end{figure}

While high-level graph optimizations can greatly improve the efficiency of DL workloads,
they are only as effective as what the operator library provides. Currently, the few DL frameworks that support operator fusion require the operator library to provide
an implementation of the fused patterns. With more network operators introduced on a regular basis, the number of possible fused kernels can grow dramatically.
This approach is no longer sustainable when targeting an increasing number of hardware back-ends since the required number of fused pattern implementations grows combinatorially with the number of data layouts, data types, and accelerator intrinsics that must be supported.
It is not feasible to handcraft operator kernels for the various operations desired by a program and for each back-end.
To this end, we next propose a code generation approach that can generate various possible implementations for a given model's operators.

\section{Generating Tensor Operations}
\label{sec:tensor}

\TensorOpt produces efficient code for each operator by generating many valid implementations on each hardware back-end and choosing an optimized implementation. This process builds on Halide's idea of decoupling descriptions
from computation rules (or \textit{schedule optimizations})~\cite{JRK:Halide} and
extends it to support new optimizations (nested parallelism, tensorization, and latency hiding) and a wide array of hardware back-ends. We now highlight \TensorOpt-specific features.

% This section addresses how \TensorOpt can provide the exploration space to
% generate fine-tuned versions of the same operator for a wide array of hardware back-ends.

\subsection{Tensor Expression and Schedule Space}
\begin{figure}[t]
\centering
\includegraphics[width=1\columnwidth]{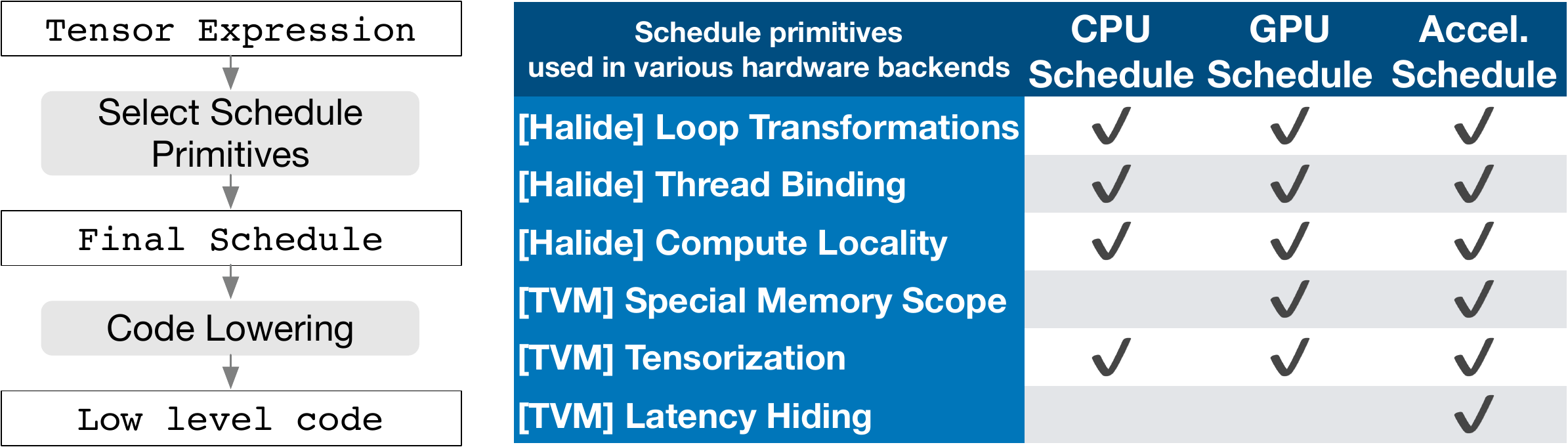}
\precap
\caption{\capsize{\TensorOpt schedule lowering and code generation process. The table lists existing Halide and novel \TensorOpt scheduling primitives being used to optimize schedules for CPUs, GPUs and accelerator back-ends.
    \REV{
    Tensorization is essential for accelerators, but it can also be used for CPUs and GPUs.
    Special memory-scope enables memory reuse in GPUs and explicit management of on-chip memory in accelerators.
    Latency hiding is specific to TPU-like accelerators.}
}}
\label{fig:codegen-flow}
\postcap
\end{figure}

We introduce a tensor expression language to support automatic code generation.
Unlike high-level computation graph representations, where the implementation of tensor operations is opaque,
each operation is described in an index formula expression language.
The following code shows an example tensor expression to compute transposed matrix multiplication:

\includegraphics[width=.98\columnwidth]{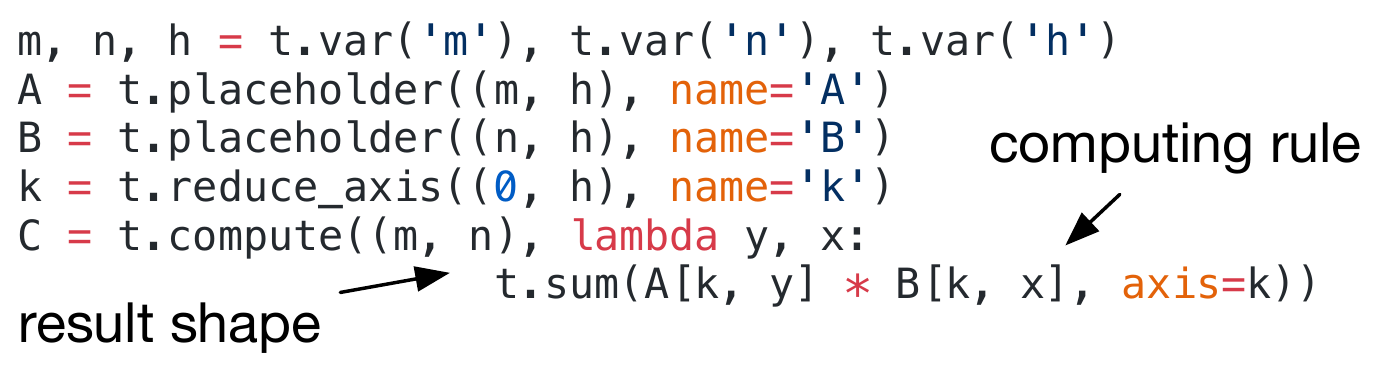}

Each compute operation specifies both the shape of the output tensor and an
expression describing how to compute each element of  it.
Our tensor expression language supports common arithmetic and math operations and covers common DL operator patterns.
The language does not specify the loop structure and many other execution details,
and it provides flexibility for adding hardware-aware optimizations for various back-ends.
Adopting the decoupled compute/schedule principle from Halide~\cite{JRK:Halide}, we use a schedule to denote a specific mapping from a tensor expression to low-level code.
Many possible schedules can perform this function.

We build a schedule by incrementally applying basic transformations~(schedule primitives) that preserve the program's logical equivalence. \autoref{fig:gemm-flow} shows an example of scheduling matrix multiplication on a specialized accelerator. Internally, \TensorOpt uses a data structure to keep track of the loop structure and other information as we apply schedule transformations. This information can then help generate low-level code for a given final schedule.

Our tensor expression takes cues from  Halide~\cite{JRK:Halide}, Darkroom~\cite{Hegarty:DarkRoom}, and TACO~\cite{Kjolstad:TACO}. Its primary enhancements include support for the new schedule optimizations discussed below.
To achieve high performance on many back-ends, we must support enough schedule primitives to cover a diverse set of optimizations on different hardware back-ends. \autoref{fig:codegen-flow} summarizes the operation code generation process and schedule primitives that \TensorOpt  supports.
We reuse helpful primitives and the low-level loop program AST from Halide, and we
introduce new primitives to optimize GPU and accelerator performance.
\REV{The new primitives are necessary to achieve optimal GPU performance and essential for accelerators.}
%\CMT{I tried to incorporate the revision into the preceding sentence.  However, the revision was unclear about whether it was accelerator performance that was also optimized.  I suggest that you therefore review the edit for technical accuracy. tqchen: accelerator performance was also optimized, but more importantly, use of accelerators are not possible without these new primitives }
CPU, GPU, TPU-like accelerators are three important types of hardware for deep learning.
This section describes new optimization primitives for CPUs, GPUs and TPU-like accelerators, while \autoref{sec:auto} explains how to automatically derive efficient schedules.

\subsection{Nested Parallelism with Cooperation}
\label{subsec:coop-paralell}

Parallelism is key to improving the efficiency of compute-intensive kernels in DL workloads.
Modern GPUs offer massive parallelism, requiring us to bake parallel patterns into schedule transformations.
Most existing solutions adopt a model called \emph{nested parallelism},
a form of fork--join. This model requires a parallel schedule primitive to parallelize a data parallel task; each task can be further recursively subdivided into subtasks to exploit the target architecture's multi-level thread hierarchy (e.g., thread groups in GPU).
We call this model \emph{shared-nothing nested parallelism} because one working thread cannot look at the data of its sibling within the same parallel computation stage.

\begin{figure}
  \centering
  \includegraphics[width=.99\linewidth]{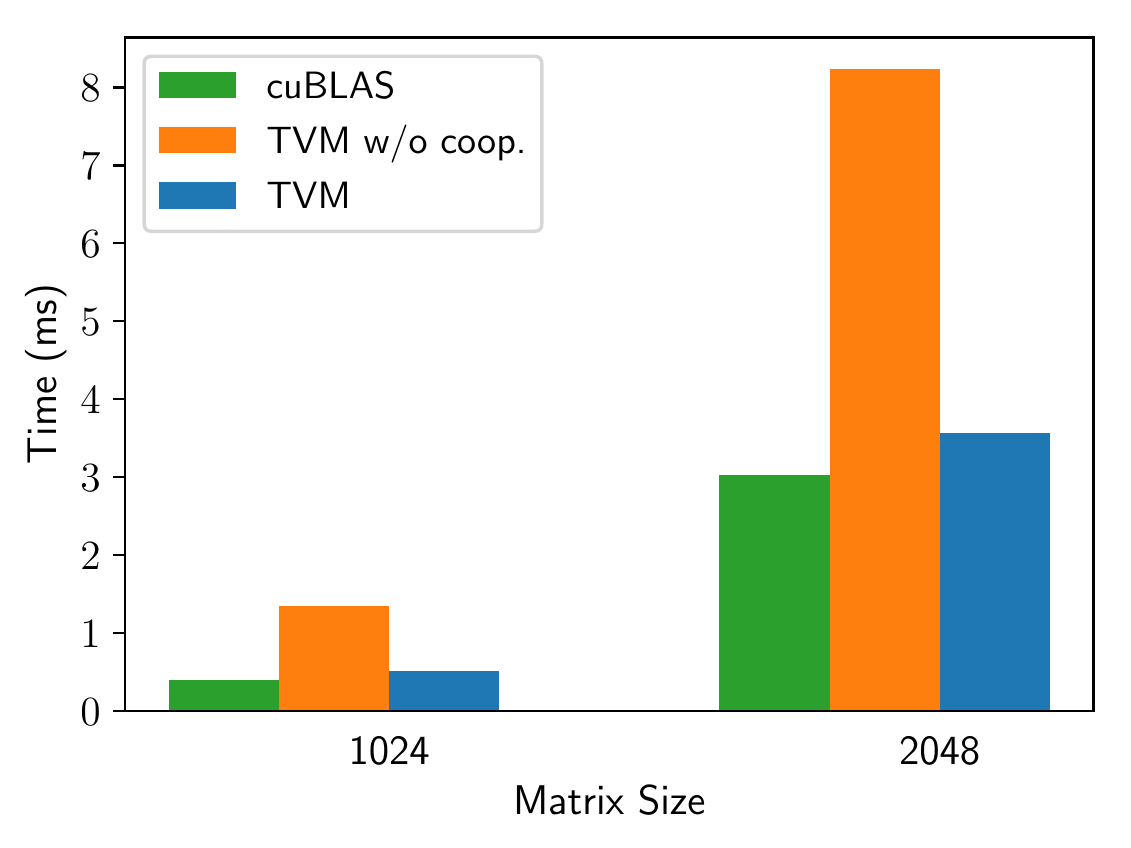}
  \caption{\REV{Performance comparison between \TensorOpt with and without cooperative shared memory fetching on matrix multiplication workloads. Tested on an NVIDIA Titan X.}}
  \label{fig:gemm-cmp}
\end{figure}

An alternative to the shared-nothing approach is to fetch data cooperatively.
Specifically, groups of threads can cooperatively fetch the data they all need and place it
%\CMT{"..and place it..."}
into a shared memory space.
 ~\footnote{
Halide recently added shared memory support but without general memory scope for accelerators.
}
This optimization can take advantage of the GPU memory hierarchy and enable data reuse across threads through shared memory regions.
 \TensorOpt supports this well-known GPU optimization using a schedule primitive \REV{to achieve optimal performance}.
The following GPU code example optimizes matrix multiplication.

\includegraphics[width=1\columnwidth]{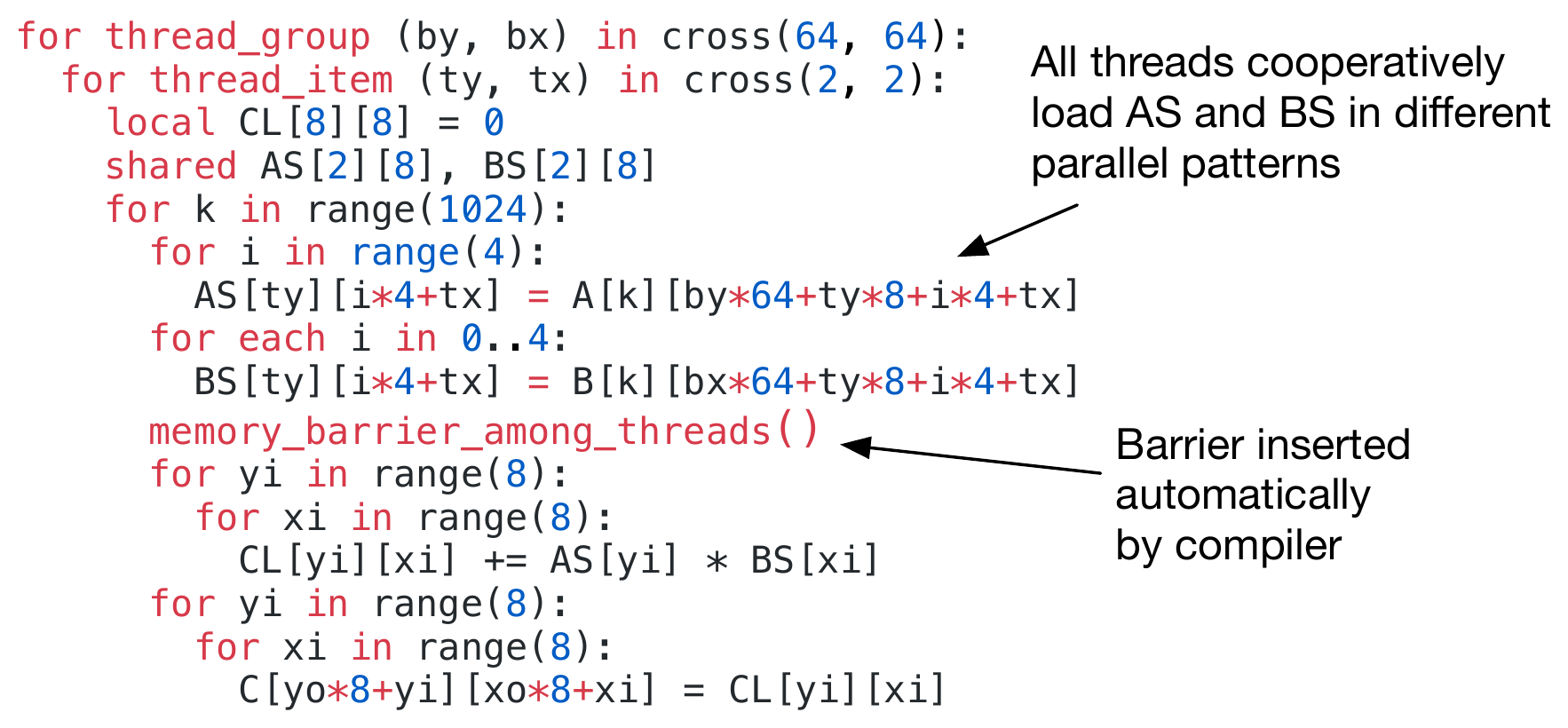}

\REV{~\autoref{fig:gemm-cmp} demonstrates the impact of this optimization.}
We introduce the concept of \emph{memory scopes}
to the schedule space so that a compute stage~(AS and BS in the code) can be marked as shared.
Without explicit memory scopes, automatic scope inference will mark compute stages
%\CMT{Can you replace "them" (unclear antecedent) with "compute stages" (assuming this is the correct referent?)}
as thread-local.
The shared task must compute the dependencies of all working threads in the group.
Additionally, memory synchronization barriers must be properly inserted to guarantee that shared loaded data is visible to consumers.
Finally, in addition to being useful to GPUs, memory scopes let us tag special memory buffers and create special lowering rules when targeting specialized DL accelerators.

\begin{figure*}[t]
\centering
\includegraphics[width=.9\textwidth]{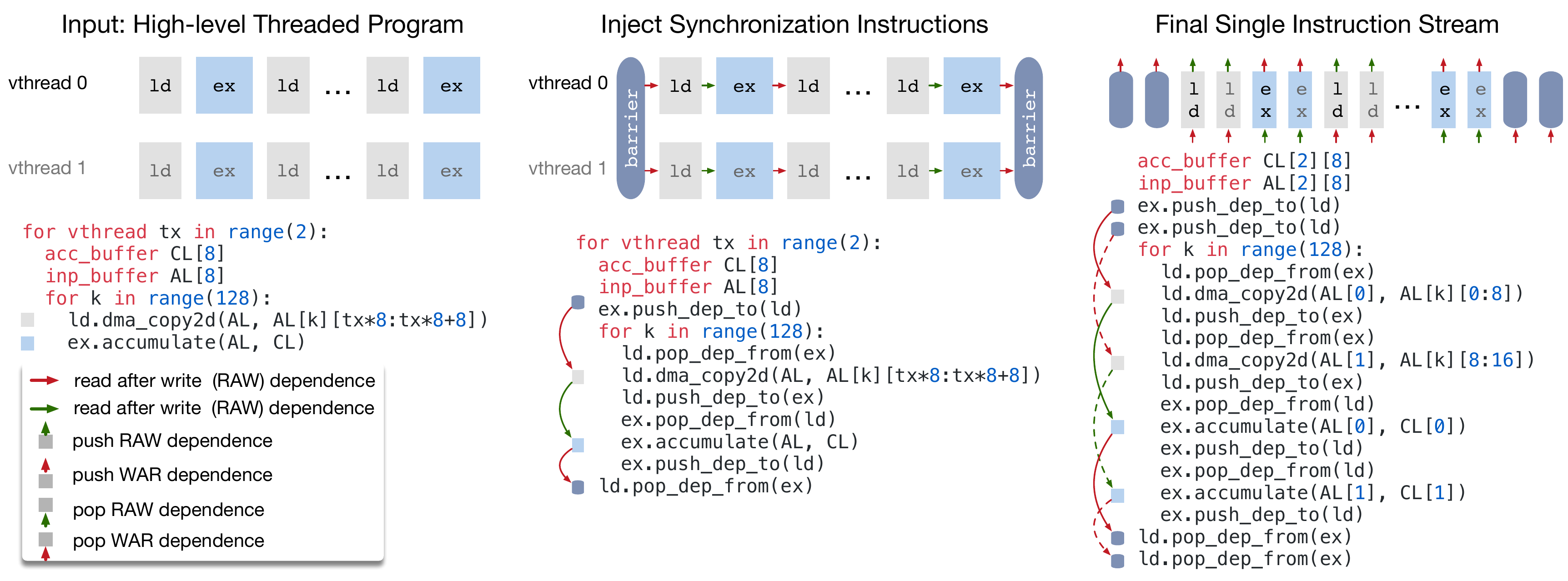}
\precap
\caption{\capsize{\TensorOpt virtual thread lowering transforms a virtual thread-parallel program to a single instruction stream; the stream contains explicit low-level synchronizations that the hardware can interpret to recover the pipeline parallelism required to hide memory access latency.}}
\postcap
\label{fig:latency_hiding}
\end{figure*}

\begin{figure}[!t]
\centering
\includegraphics[width=.9\columnwidth]{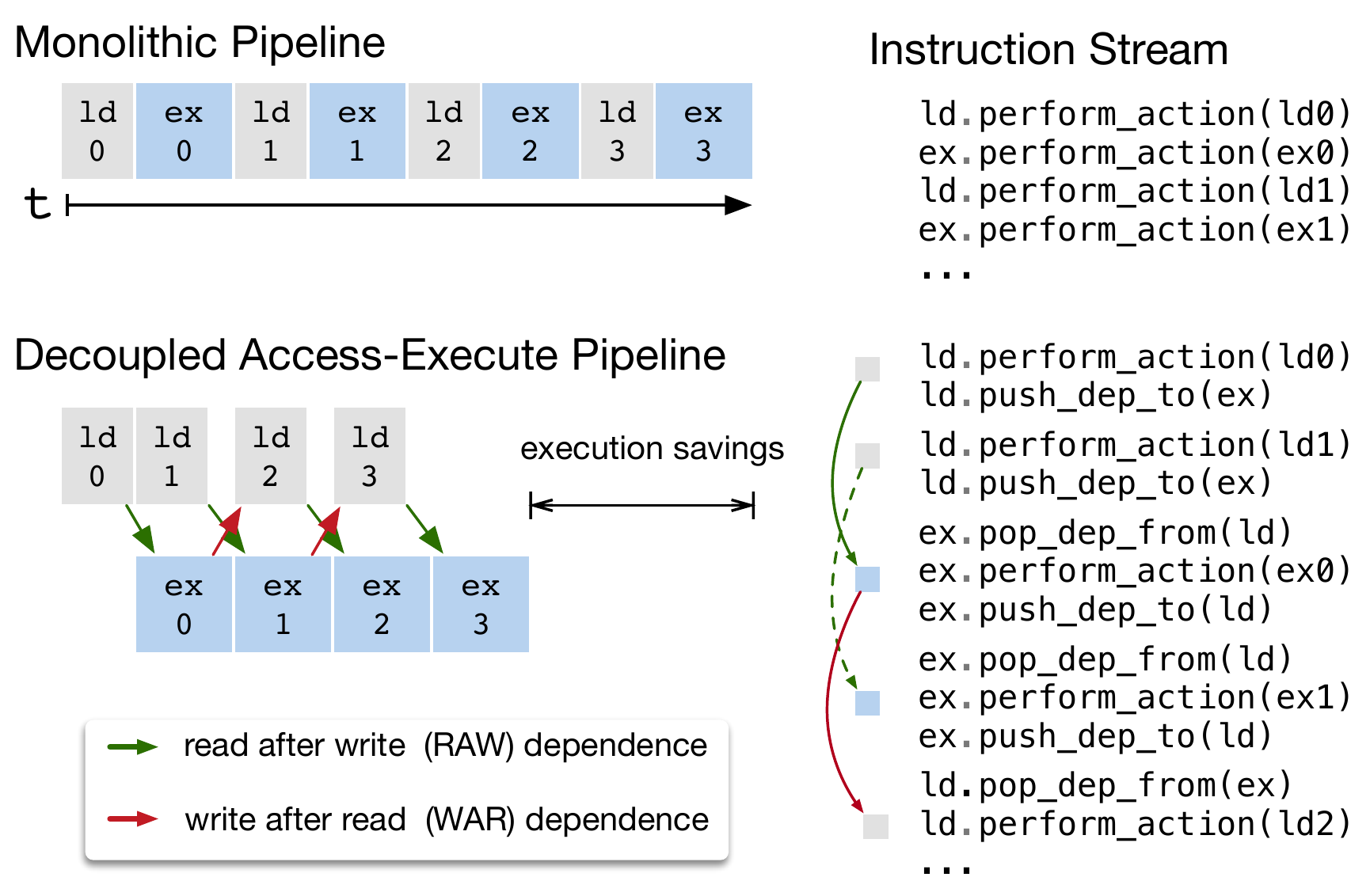}
\precap
\caption{\capsize{Decoupled Access-Execute in hardware hides most memory access latency by allowing memory and computation to overlap. Execution correctness is enforced by low-level synchronization in the form of dependence token enqueueing/dequeuing actions, which the compiler stack must insert in the instruction stream.}}
\postcap
\label{fig:decoupled_access_execute}
\end{figure}

\subsection{Tensorization}

DL workloads have high arithmetic intensity, which can typically be decomposed into tensor operators like
matrix-matrix multiplication or 1D convolution. These natural decompositions have led to the recent trend of adding tensor compute primitives
~\cite{Jouppi:TPU,volta-whitepaper,Chen:Eyeriss}.
These new primitives create both opportunities and challenges for schedule-based compilation; while using them can improve performance, the compilation framework must seamlessly integrate them.
We dub this \emph{tensorization}: it is analogous to vectorization for SIMD architectures but has significant differences. Instruction inputs are multi-dimensional, with fixed or variable lengths, and each has different data layouts.
More importantly, we cannot support a fixed set of primitives since new accelerators are emerging
with their own variations of tensor instructions. We therefore need an \emph{extensible} solution.

We make tensorization extensible by separating the target hardware intrinsic from the schedule with a mechanism for  tensor-intrinsic declaration.
We use the same tensor expression language to declare both the
behavior of each new hardware intrinsic and the lowering rule associated with it.
The following code shows how to declare an $8\times8$ tensor hardware intrinsic.

\includegraphics[width=\columnwidth]{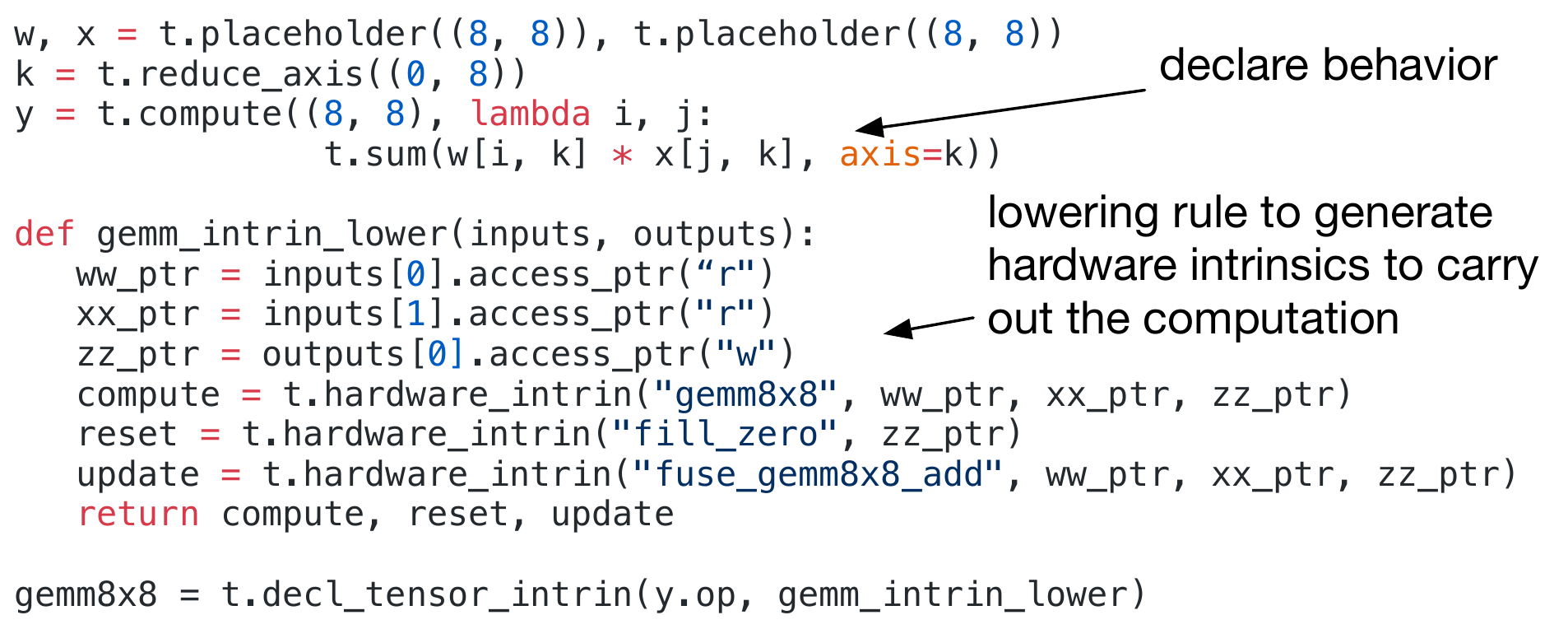}

Additionally, we introduce a \emph{tensorize} schedule primitive to replace a
unit of computation with the corresponding intrinsics.
The compiler matches the computation pattern with a hardware declaration and lowers it to the corresponding  hardware intrinsic.

Tensorization decouples the schedule from specific hardware primitives,
making it easy to extend \TensorOpt to support new hardware architectures.
The generated code of tensorized schedules aligns with  practices in high-performance computing: break complex operations into a sequence of micro-kernel calls.
We can also use the \emph{tensorize} primitive to take advantage
of handcrafted micro-kernels, which can be beneficial in some
platforms.
For example, we implement ultra low precision operators for mobile CPUs that operate on data types that are one- or two-bits wide by leveraging a bit-serial matrix vector multiplication micro-kernel. This micro-kernel accumulates results into progressively larger data types to minimize the memory footprint.
Presenting the micro-kernel as a tensor intrinsic to \TensorOpt yields up to a 1.5$\times$ speedup over the non-tensorized version.

% \begin{figure}[t]
% \centering
% \includegraphics[width=.8\columnwidth]{sync_program_model}
% \precap
% \caption{\small{Low-level program with explicit synchronization.
%          This example demonstrates how to compute $C_j = \sum_k A_{k,j}$.
%          The hardware contains two instruction streams that need to be explicitly synchronized.
%          Carefully crafting the messages enables execution overlap between instruction streams.} \TODO{should we call q0 access and q1 execute, in the spirit of the access/execute model?}}
% \postcap
% \label{fig:sync_program_model}
% \end{figure}

\subsection{Explicit Memory Latency Hiding}
\label{sec:smt}

\textit{Latency hiding} refers to the process of overlapping memory operations with computation to maximize utilization of memory and compute resources.
It requires different strategies depending on the target hardware back-end.
On CPUs, memory latency hiding is achieved implicitly with simultaneous multithreading~\cite{Eggers:SMT} or hardware prefetching~\cite{Jouppi:CachePerf, Chen:Prefetching}.
GPUs rely on rapid context switching of many warps of threads~\cite{Volkov:GPUs}.
In contrast, specialized DL accelerators such as the TPU~\cite{Jouppi:TPU} usually favor leaner control with a \emph{decoupled access-execute} (DAE) architecture~\cite{Smith:DAE} and offload the problem of fine-grained synchronization to software.

\autoref{fig:decoupled_access_execute} shows a DAE hardware pipeline that reduces runtime latency.
Compared to a monolithic hardware design, the pipeline can hide most memory access overheads and almost fully utilize compute resources.
To achieve higher utilization, the instruction stream must be augmented with fine-grained synchronization operations. Without them, dependencies cannot be enforced, leading to erroneous execution.
Consequently, DAE hardware pipelines require fine-grained dependence enqueuing/dequeuing operations between the pipeline stages
%\CMT{..."between pipeline stage\textbf{s}"? "Between" what and what?}
to guarantee correct execution, as shown in \autoref{fig:decoupled_access_execute}'s instruction stream.

Programming DAE accelerators that require explicit low-level synchronization is difficult.
To reduce the programming burden, we introduce a virtual threading scheduling primitive that lets programmers specify a high-level data parallel program as they would a hardware back-end with support for multithreading.
\TensorOpt then automatically lowers the program to a single instruction stream with low-level explicit synchronization, as shown in \autoref{fig:latency_hiding}.
The algorithm starts with a high-level multi-threaded program schedule and then inserts the necessary low-level synchronization operations to guarantee correct execution within each thread.
Next, it
%\CMT{[passive voice was used here.  Does change to active voice and use of "it" correctly refer to "the algorithm"?]}
interleaves operations of all virtual threads into a single instruction stream.
Finally, the hardware recovers the available pipeline parallelism dictated by the low-level synchronizations in the instruction stream.

\paragraph{Hardware Evaluation of Latency Hiding.}
\begin{figure}[!t]
\centering
\includegraphics[width=.95\columnwidth]{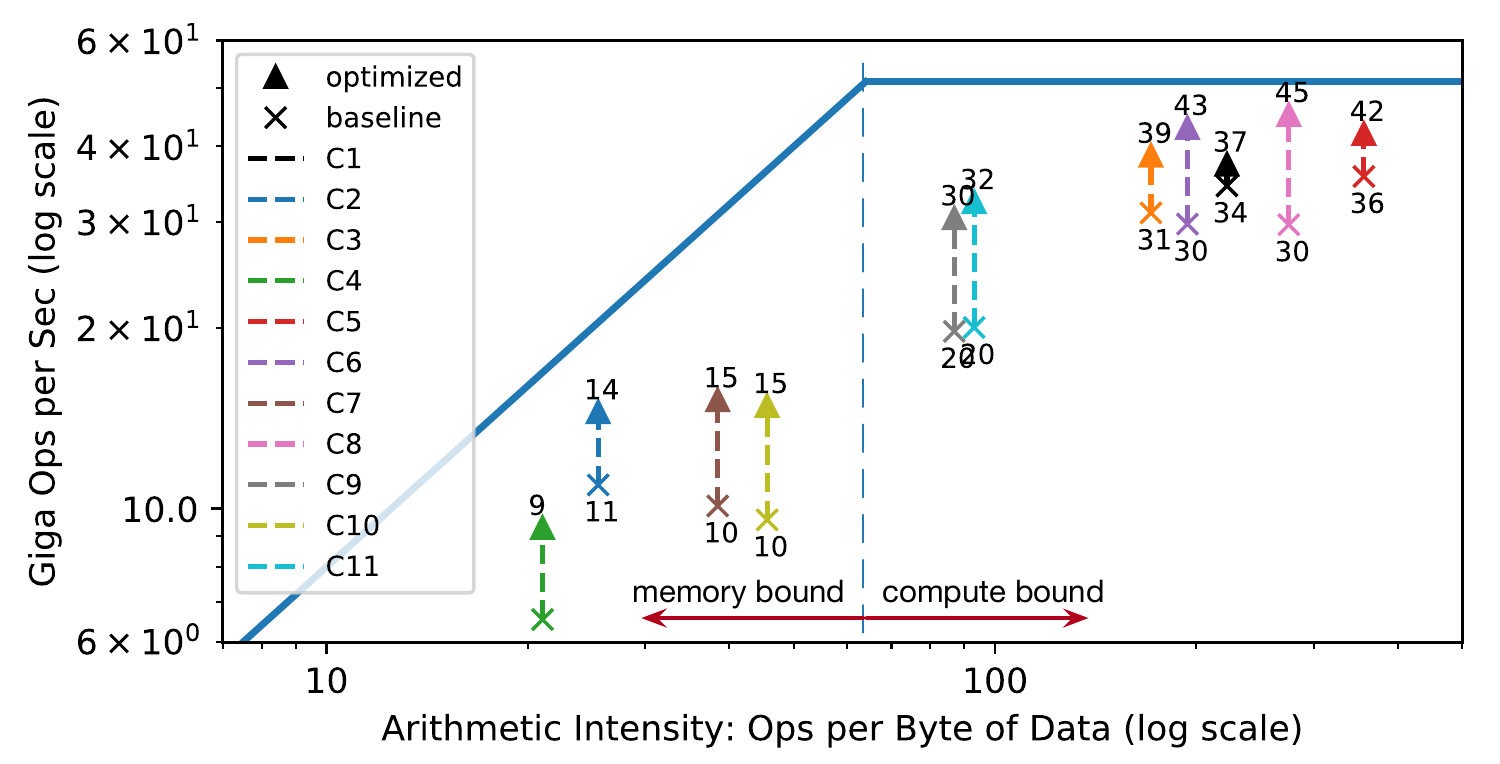}
\precap
\caption{\capsize{Roofline~\cite{roofline} of an FPGA-based DL accelerator running ResNet inference. With latency hiding enabled by \TensorOpt, performance of the benchmarks is brought closer to the roofline, demonstrating higher compute and memory bandwidth efficiency.}}
\postcap
\label{fig:fpga_latencyhiding}
\end{figure}

We now demonstrate the effectiveness of latency hiding on a custom FPGA-based accelerator design, which we describe in  depth in \autoref{sec:fpga_eval}.
We ran each layer of ResNet on the accelerator and used \TensorOpt to generate two schedules: one with latency hiding, and one without.
The schedule with latency hiding parallelized the program with virtuals threads
%\CMT{"virtual threads"?}
to expose pipeline parallelism and therefore hide memory access latency.
Results are shown in \autoref{fig:fpga_latencyhiding} as a roofline diagram~\cite{roofline};
roofline performance diagrams provide insight into how well a given system uses computation and memory resources for different benchmarks.
Overall, latency hiding improved performance on all ResNet layers.
Peak compute utilization increased from 70\% with no latency hiding to 88\% with latency hiding.
% These results demonstrate \TensorOpt's ability to take advantage of modern deep learning accelerators.

\section{Automating Optimization}
\label{sec:auto}

\begin{figure}[t]
\centering
\includegraphics[width=1\columnwidth]{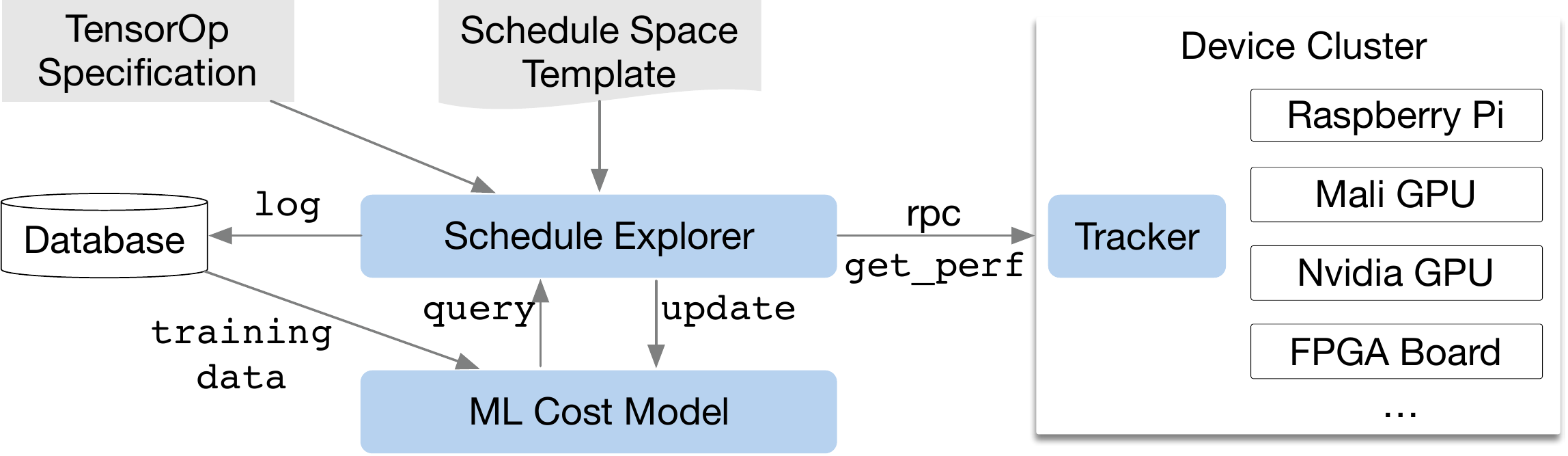}
\precap
\caption{\capsize{Overview of automated optimization framework.
    A schedule explorer examines the schedule space using an ML-based cost model and chooses
    experiments to
    run on a distributed device cluster via RPC.
    To improve its predictive power, the ML model is updated periodically using
    collected data recorded in a database.
}}
\postcap
\label{fig:autoframework}
\end{figure}

Given the rich set of schedule primitives, our remaining problem is to
find optimal operator implementations for each layer of a DL model. Here,
\TensorOpt creates a specialized operator for the specific input shape and layout associated with each layer.
Such specialization offers significant performance benefits (in contrast to handcrafted code that would target a smaller diversity of shapes and layouts), but it also raises automation challenges.
The system needs to choose the schedule optimizations -- such as modifying the loop order or optimizing for the memory hierarchy -- as well as schedule-specific parameters, such as the tiling size and the loop unrolling factor. Such combinatorial choices create a large search space of operator implementations for each hardware back-end.  To address this challenge, we built an \textit{automated schedule optimizer} with two main components: a schedule explorer that \emph{proposes} promising new configurations, and a machine learning cost model that \emph{predicts} the performance of a given configuration. This section describes these components and \TensorOpt's automated optimization flow~(\autoref{fig:autoframework}).

\subsection{Schedule Space Specification}
We built a \textit{schedule template specification API} to let a developer declare knobs in the schedule space.
The template specification allows incorporation of a developer's domain-specific knowledge, as necessary, when specifying possible schedules.
We also created a \textit{generic master template for each hardware back-end} that automatically extracts possible
knobs based on the computation description expressed using the tensor expression language.
At a high level, we would like to consider as many configurations as possible and let the optimizer manage the selection burden.
Consequently, the optimizer must search over \emph{billions} of possible configurations for the real world DL workloads used in our experiments.

\subsection{ML-Based Cost Model}

One way to find the best schedule from a large configuration space is through blackbox optimization, i.e., auto-tuning.
This method is used to tune high performance computing libraries\cite{Whaley:ATLAS,FFTW}.
However, auto-tuning requires many experiments to identify a good configuration.

An alternate approach is to build a predefined cost model to guide the search for a particular hardware back-end instead of running all possibilities and measuring their performance.
Ideally, a perfect cost model considers all factors affecting performance: memory access patterns, data reuse, pipeline dependencies, and threading patterns, among others.  This approach, unfortunately, is burdensome due to the increasing complexity of modern hardware.
Furthermore, every new hardware target requires a new (predefined) cost model.

\vspace{1em}

\begin{table}
\centering
\begin{footnotesize}
\begin{tabular}{lcccc}
\hline
\parbox[l]{.27 \linewidth}{Method Category} &
\parbox[c]{.06 \linewidth}{Data Cost} &
\parbox[c]{.07 \linewidth}{Model Bias} &
\parbox[c]{.14 \linewidth}{Need Hardware Info}&
\parbox[c]{.10 \linewidth}{Learn from History} \\
\hline
Blackbox auto-tuning & high & none & no & no\\
Predefined cost model & none & high & yes & no\\
\textbf{ML based cost model} & \textbf{low} & \textbf{low} & \textbf{no} & \textbf{yes}\\
\hline
\end{tabular}
\end{footnotesize}
\precap
\caption{\capsize{
    Comparison of automation methods.
    Model bias refers to inaccuracy due to modeling.
}
}
\label{tbl:autocmp}
\postcap
\end{table}
We instead take a statistical approach to solve the cost modeling problem.
In this approach, a schedule explorer proposes configurations that may improve an operator's performance.
For each schedule configuration, we use an ML model that takes the lowered loop program as input and predicts its running time on a given hardware back-end.
The model, trained using runtime measurement data collected during exploration, does not require the user to input detailed hardware information.
We update the model periodically as we explore more configurations during optimization, which  improves accuracy for other related workloads, as well.
In this way, the quality of the ML model improves with more experimental trials.
~\autoref{tbl:autocmp} summarizes the key differences between automation methods.
ML-based cost models strike a balance between auto-tuning and predefined cost modeling
and can benefit from the historical performance data of related workloads.

\paragraph{Machine Learning Model Design Choices.}

\begin{figure}
	\centering
	\includegraphics[width=\columnwidth]{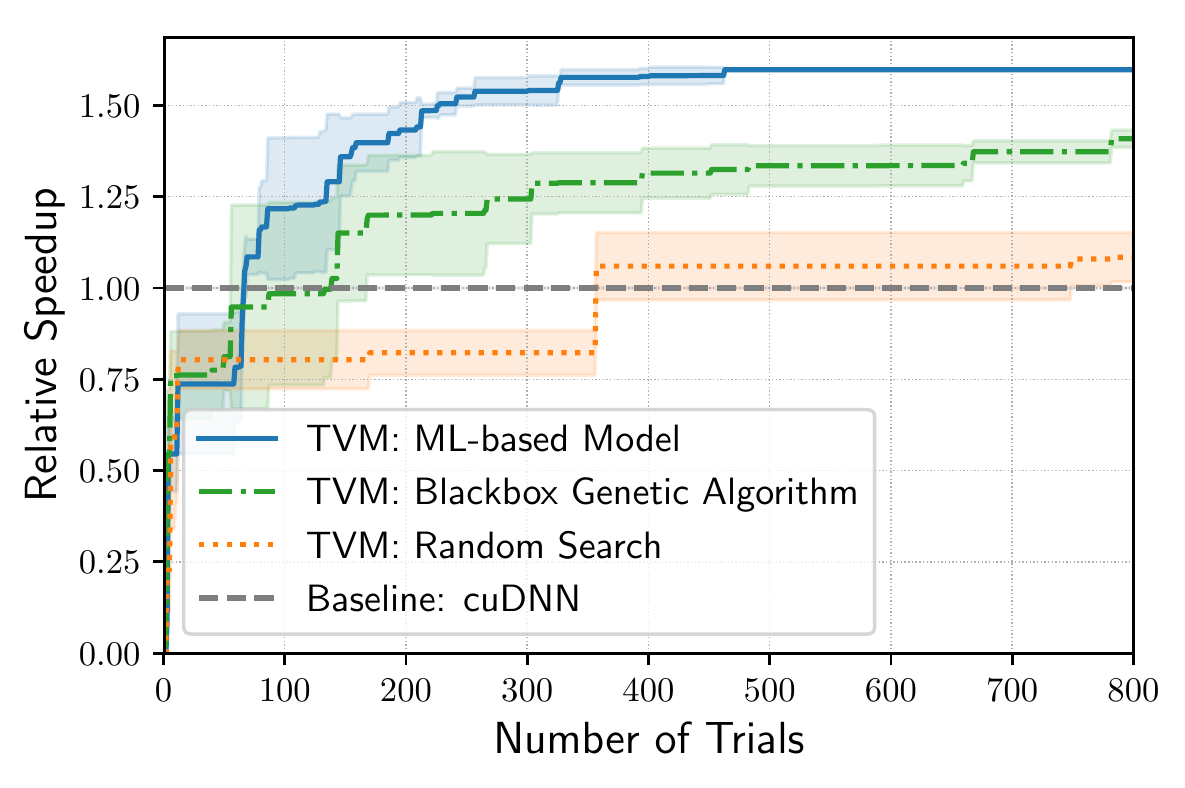}
	\precap
	\caption{Comparison of different automation methods for a conv2d operator in ResNet-18 on TITAN X.
    The ML-based model starts with no training data and uses the collected data to improve itself.
    The Y-axis is the speedup relative to cuDNN. We observe a similar trend for other workloads.}
	\postcap
	\label{fig:tuners}
\end{figure}

We must consider two key factors when choosing which ML model the schedule explorer will use: \emph{quality} and \emph{speed}.
The schedule explorer queries the cost model frequently, which incurs overheads due to model prediction time and model refitting time.
To be useful, these overheads must be smaller than the time it takes to measure performance on real hardware, which can be on the order of seconds depending on the specific workload/hardware target.
This speed requirement differentiates our problem from traditional hyperparameter tuning problems, where the cost of performing measurements is very high relative to model overheads, and more expensive models can be used.
In addition to the choice of model, we need to choose an objective function to train the model, such as the error in a configuration's predicted running time.
However, since the explorer selects the top candidates based only on the relative order of the prediction~(A runs faster than B),
we need not predict \REV{the absolute} execution times directly.
Instead, we use a rank objective to predict the relative order of runtime costs.

\begin{figure}[t]
	\centering
	\includegraphics[width=\columnwidth]{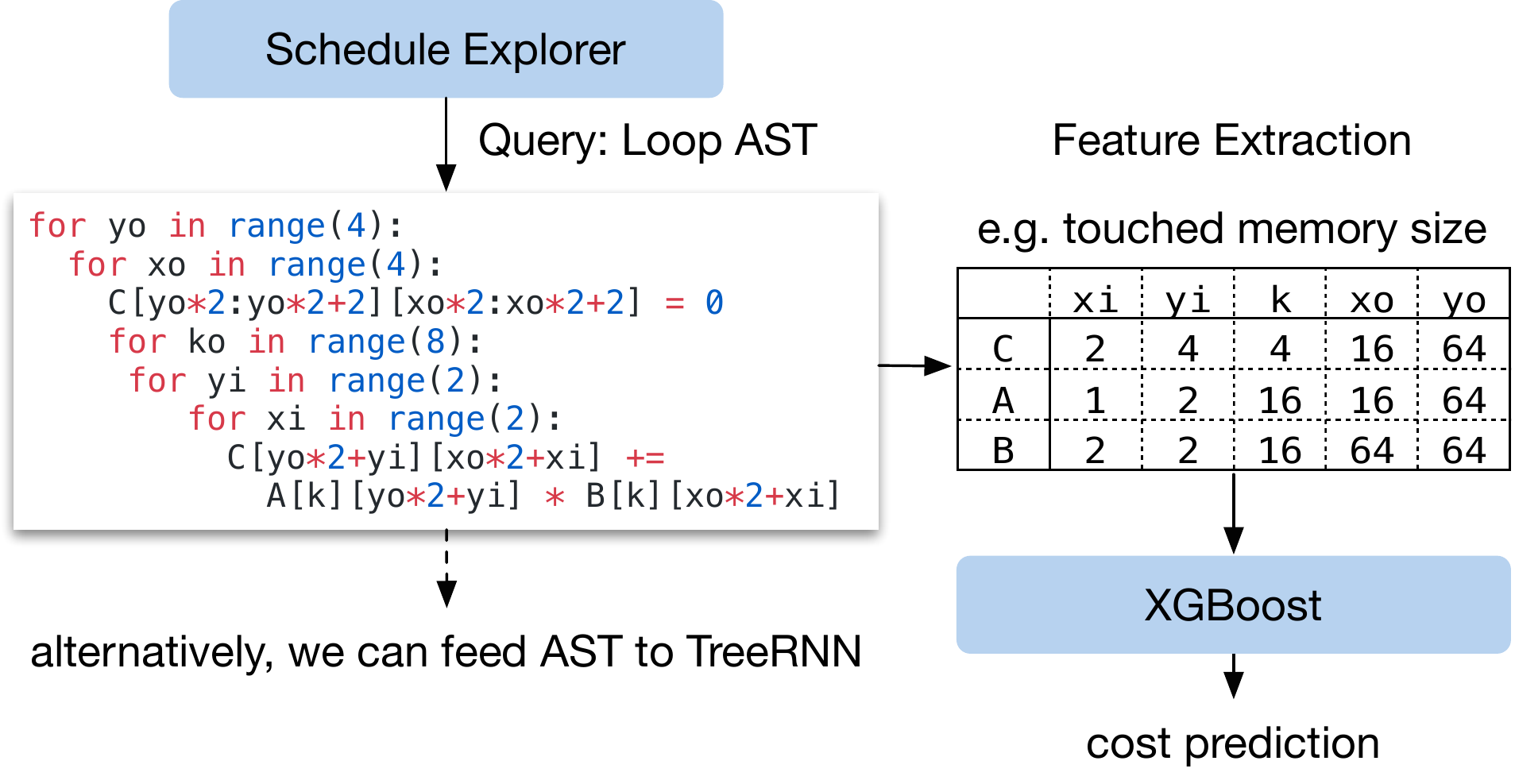}
	\precap
	\caption{\capsize{Example workflow of ML cost models. XGBoost predicts costs based on loop program features. TreeRNN directly summarizes the AST.}}
	\postcap
	\label{fig:ml_flow}
\end{figure}

We implement several types of models in our ML optimizer. We employ a \textit{gradient tree boosting model}~(based on XGBoost~\cite{XGBoostKDD}), which makes predictions based on features extracted from the loop program;
these features include the memory access count and reuse ratio of each memory buffer at each loop level, as well as a one-hot encoding of loop annotations such as ``vectorize'', ``unroll'', and ``parallel.''
We also evaluate a \textit{neural network model} that uses TreeRNN~\cite{TreeRNN} to summarize the loop program's AST without feature engineering. \autoref{fig:ml_flow} summarizes the workflow of the cost models.
We found that tree boosting and TreeRNN have similar predictive quality.
However, the former performs prediction twice as fast and costs much less time to train. As a result, we chose gradient tree boosting as the default cost model in our experiments.
Nevertheless, we believe that both approaches are valuable and expect more future research on this problem.

On average, the tree boosting model does prediction in 0.67 ms, thousands of times faster than running a real measurement. \autoref{fig:tuners} compares an ML-based optimizer to blackbox auto-tuning methods; the former finds better configurations much faster than the latter.

%\begin{figure}[t]
%	\centering
%	\includegraphics[width=\columnwidth]{tree_rnn}
%	\precap
%	\caption{The structure of a TreeRNN for matrix multiplication. We use the simplified AST of lowered IR as tree structure and GRU Cell as node. Each cell takes the sum of its children as hidden state. We train an embedding vector for every node and feed in the embedding, length of for loop and type of for loop as input. No feature engineering is required.}
%	\postcap
%	\label{fig:tree_rnn}
%\end{figure}

\subsection{Schedule Exploration}
Once we choose a cost model, we can use it to select promising configurations on which to iteratively run real measurements.
In each iteration, the explorer uses the ML model's predictions to select a batch of candidates on which to run the measurements.
The collected data is then used as training data to update the model.
If no initial training data exists, the explorer picks random candidates to measure.

The simplest exploration algorithm enumerates and runs every configuration through the cost model, selecting the top-$k$ predicted performers.
However, this strategy becomes intractable with large search spaces.
Instead, we run a parallel simulated annealing algorithm~\cite{SimulatedAnealing}.
The explorer starts with random configurations, and, at each step, randomly walks to a nearby configuration.
This transition is successful if cost decreases as predicted by the cost model. It is likely to
fail~(reject) if the target configuration has a higher cost.
The random walk tends to converge on configurations that have lower costs as predicted by the cost model.
Exploration states persist across cost model updates; we continue from the last configuration after these updates.

% removed the schedule space example table because we didn't explain the knobs
% A schedule space example is shown in \autoref{tbl:schedule_space}
%\begin{figure}
%\centering
%\includegraphics[width=.9\columnwidth]{rpc}
%\precap
%\caption{RPC support in \TensorOpt runtime to make it easy to cross compile,
%         deploy, and profile remote devices.}
%\postcap
%\label{fig:rpc}
%\end{figure}

\subsection{Distributed Device Pool and RPC}

A \textit{distributed device pool} scales up the running of on-hardware trials and enables fine-grained resource sharing among multiple optimization jobs.
\TensorOpt implements a customized, RPC-based distributed device pool that enables clients to run programs on a specific type of device.
We can use this interface to compile a program on the host compiler, request a remote device,
run the function remotely, and access results in the same script on the host.
\TensorOpt's RPC supports dynamic upload and runs cross-compiled modules and functions that use its runtime convention.
As a result, the same infrastructure can perform a single workload optimization and end-to-end graph inference.
Our approach automates the compile, run, and profile steps across multiple devices.
This infrastructure is especially critical for embedded devices,
which traditionally require tedious manual effort for cross-compilation, code deployment, and measurement.

\section{Evaluation}
\label{sec:exp}
% \begin{table}[t]
% 	\begin{tabular}{ll}
% 	\hline
%   Module being Evaluated & Location\\
% 	\hline
%   Operator fusion & \autoref{fig:fusion-cmp} in \autoref{sec:graph}\\
%   Latency hiding & \autoref{fig:fpga_latencyhiding} in \autoref{sec:tensor}\\
%   ML-based cost model & \autoref{fig:tuners} in \autoref{sec:auto}\\
% 	\hline
% 	\end{tabular}
% 	\centering
% 	\precap
% 	\caption{\small{Location of Experimental results in previous sections to evaluate
%       the impact of individual optimization.}}
% 	\postcap
% 	\label{tbl:prev-exp}
% \end{table}

\begin{table}[t]
  \begin{footnotesize}
	\begin{tabular}{ccccc}
	\hline
	Name & Operator & $H, W$ & $IC, OC$ & $K, S$ \\
	\hline
	C1  & conv2d & 224, 224 & 3,64  & 7, 2 \\
	C2  & conv2d & 56, 56 & 64,64   & 3, 1 \\
	C3  & conv2d & 56, 56 & 64,64   & 1, 1 \\
	C4  & conv2d & 56, 56 & 64,128  & 3, 2 \\
	C5  & conv2d & 56, 56 & 64,128  & 1, 2 \\
	C6  & conv2d & 28, 28 & 128,128 & 3, 1 \\
	C7  & conv2d & 28, 28 & 128,256 & 3, 2 \\
	C8  & conv2d & 28, 28 & 128,256 & 1, 2 \\
	C9  & conv2d & 14, 14 & 256,256 & 3, 1 \\
	C10 & conv2d & 14, 14 & 256,512 & 3, 2 \\
	C11 & conv2d & 14, 14 & 256,512 & 1, 2 \\
	C12 & conv2d &  7,  7 & 512,512 & 3, 1 \\
	\hline
	\end{tabular}
	\begin{tabular}{ccccc}
		\hline
		Name & Operator & $H, W$ & $IC$ & $K, S$ \\
		\hline
		D1  & depthwise conv2d & 112, 112 & 32 & 3, 1 \\
		D2  & depthwise conv2d & 112, 112 & 64 & 3, 2 \\
		D3  & depthwise conv2d & 56, 56 & 128 & 3, 1 \\
		D4  & depthwise conv2d & 56, 56 & 128 & 3, 2 \\
		D5  & depthwise conv2d & 28, 28 & 256 & 3, 1 \\
		D6  & depthwise conv2d & 28, 28 & 256 & 3, 2 \\
		D7  & depthwise conv2d & 14, 14 & 512 & 3, 1 \\
		D8  & depthwise conv2d & 14, 14 & 512 & 3, 2 \\
		D9  & depthwise conv2d &  7,  7 & 1024& 3, 1 \\
		\hline
	\end{tabular}
  \end{footnotesize}
	\centering
	\precap
	\caption{\capsize{Configurations of all conv2d operators in ResNet-18 and all depthwise conv2d operators in MobileNet used in the single kernel experiments.
	H/W denotes height and width, IC input channels, OC output channels,
	K kernel size, and S stride size. All ops use ``SAME'' padding. All depthwise conv2d operations have channel multipliers of 1.}}
	\postcap
	\label{tbl:all-op}
\end{table}

\TensorOpt's core is implemented in C++~($\sim$50k LoC).
We provide language bindings to Python and Java.
Earlier sections of this paper evaluated the impact of several individual optimizations and components of \TensorOpt, namely, \textit{operator fusion} in \autoref{fig:fusion-cmp},   \textit{latency hiding} in \autoref{fig:fpga_latencyhiding}, and
the \textit{ML-based cost model} in \autoref{fig:tuners}.
We now focus on an end-to-end evaluation that aims to  answer the following questions:
\begin{itemize}[noitemsep]
  \item Can \TensorOpt optimize DL workloads over multiple platforms?
  \item How does \TensorOpt compare to existing DL frameworks~(which rely on heavily optimized libraries) on each back-end?
  \item Can \TensorOpt support new, emerging DL workloads (e.g., depthwise convolution, low precision operations)?
  \item Can \TensorOpt support and optimize for new specialized accelerators?
\end{itemize}

To answer these questions, we evaluated \TensorOpt on four types of platforms: (1) a server-class GPU,  (2) an embedded GPU, (3) an embedded CPU, and (4) a DL accelerator implemented on a low-power FPGA SoC.
The benchmarks are based on real world DL inference workloads, including ResNet~\cite{He2016}, MobileNet~\cite{Howard:MobileNet}, the LSTM Language Model~\cite{zaremba2014recurrent}, the Deep Q Network (DQN)~\cite{mnih2015human} and Deep Convolutional Generative Adversarial Networks (DCGAN)~\cite{radford2015dcgan}. We compare our approach to existing DL frameworks, including MxNet~\cite{MXNet-whitepaper} and TensorFlow~\cite{tensorflow2015-whitepaper}, that rely on highly engineered, vendor-specific libraries. \TensorOpt performs end-to-end automatic optimization and code generation
\emph{without the need for an external operator library}.

\subsection{Server-Class GPU Evaluation}
\begin{figure}[t]
 \centering
 \includegraphics[width=\columnwidth]{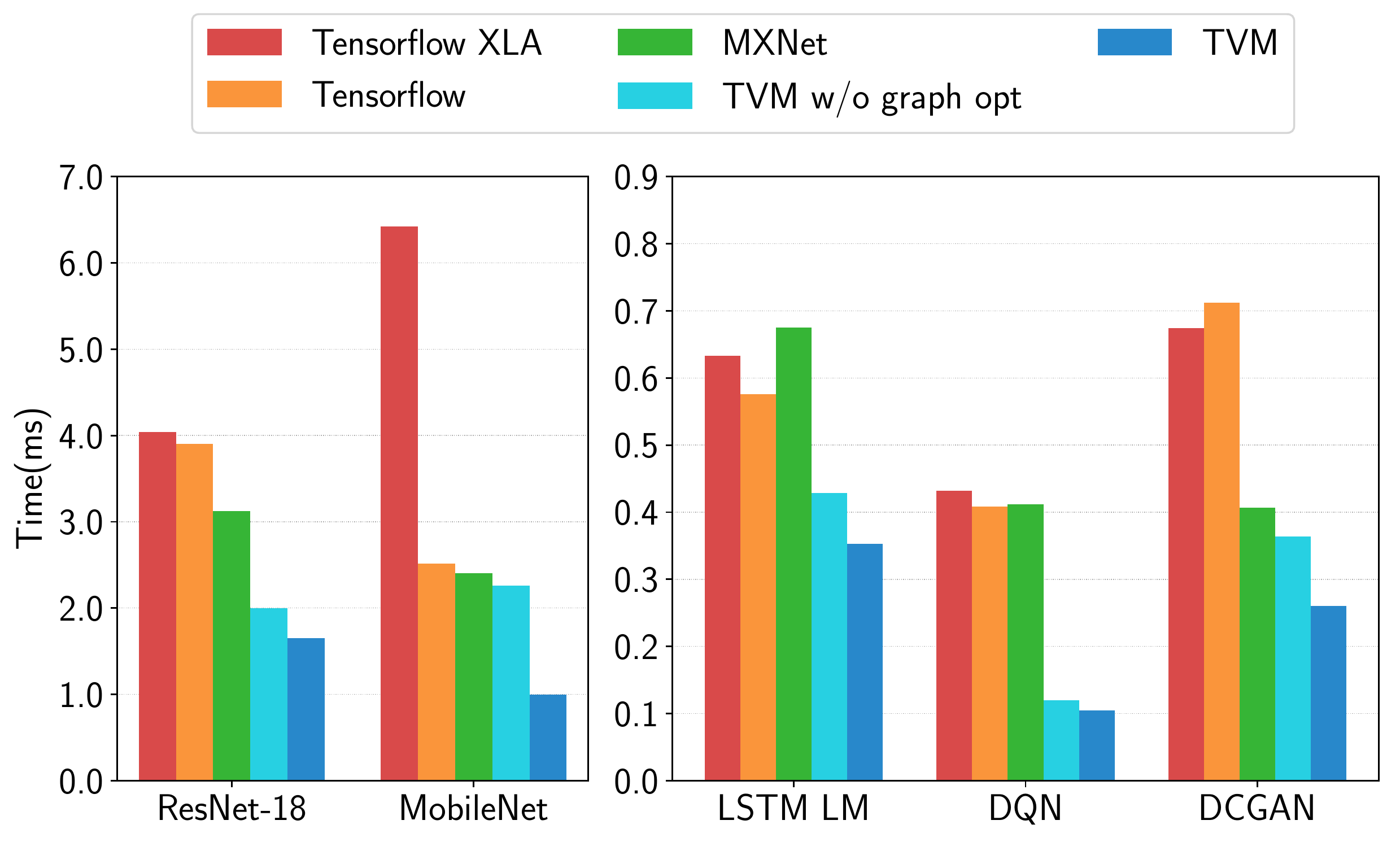}
 \precap
 \caption{GPU end-to-end evaluation for \TensorOpt, MXNet, Tensorflow, and Tensorflow XLA. Tested on the NVIDIA Titan X.}
 \postcap
 \label{fig:cuda-end2end}
\end{figure}

We first compared the end-to-end performance of deep neural networks \TensorOpt, MXNet~(v1.1), Tensorflow~(v1.7), and Tensorflow XLA on an Nvidia Titan X.  MXNet and Tensorflow both use cuDNN v7 for convolution operators; they implement their own versions of depthwise convolution since it is relatively new and not yet supported by the latest libraries. They also use cuBLAS v8 for matrix multiplications.
On the other hand, Tensorflow XLA uses JIT compilation.

\autoref{fig:cuda-end2end} shows that \TensorOpt outperforms the baselines, with speedups ranging from 1.6$\times$ to 3.8$\times$ due to both joint graph optimization
and the automatic optimizer, which generates high-performance fused operators.
DQN's 3.8 x speedup results from its use
of unconventional operators~(4$\times$4 conv2d, strides=2) that are not well optimized by cuDNN; the ResNet workloads are more conventional.
\TensorOpt automatically finds optimized operators in both cases.

\begin{figure}[t]
	\centering
	\includegraphics[width=\columnwidth]{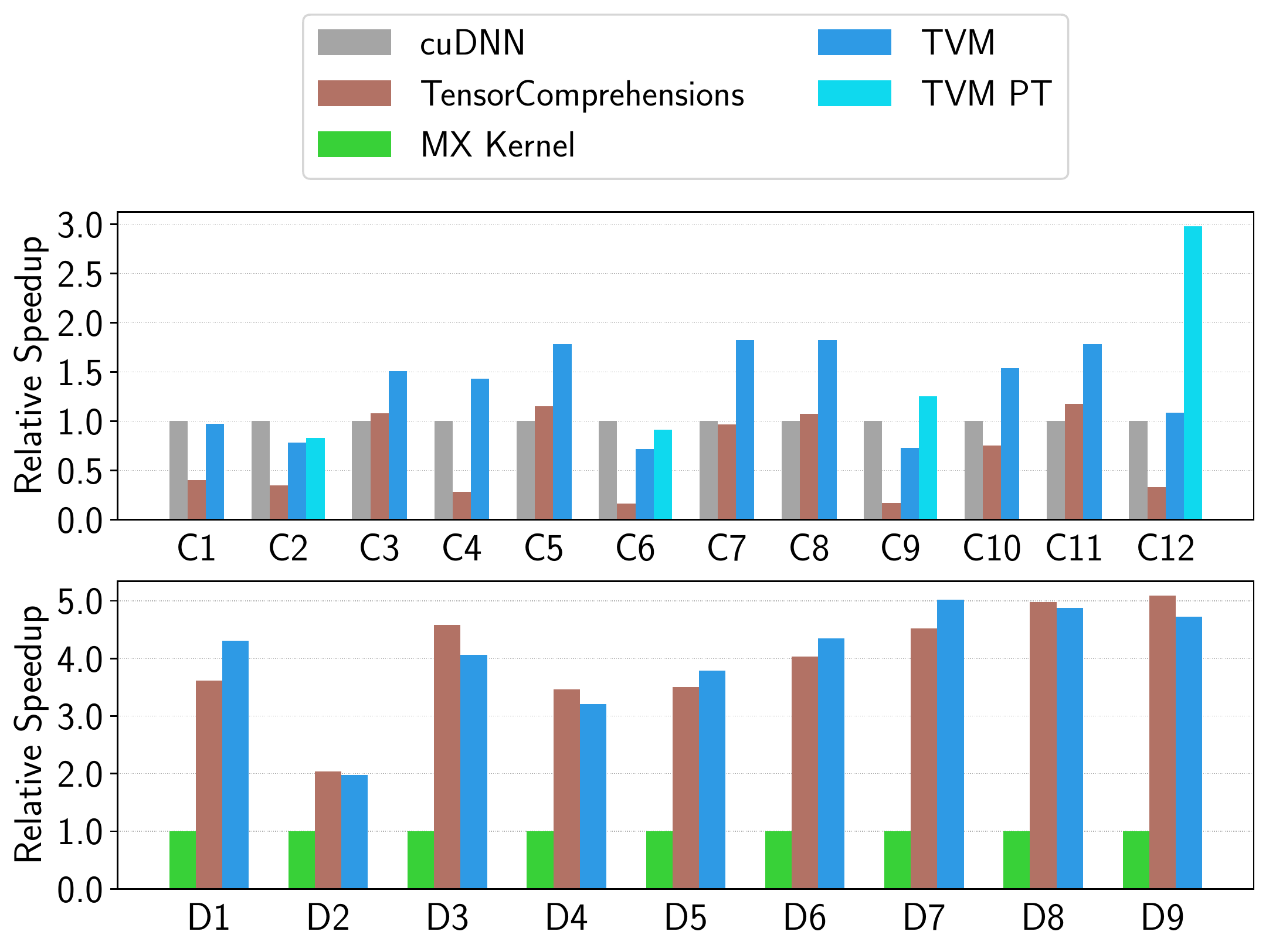}
	\precap
	\caption{\capsize{Relative speedup of all conv2d operators in ResNet-18 and all depthwise conv2d operators in MobileNet. Tested on a TITAN X. See \autoref{tbl:all-op} for operator configurations.
     We also include a weight pre-transformed Winograd~\cite{Winograd} for 3x3 conv2d~(\TensorOpt PT).
    }}
	\postcap
	\label{fig:cuda-op}
\end{figure}

To evaluate the effectiveness of operator level optimization,
we also perform a breakdown comparison for each tensor operator in ResNet and MobileNet, shown in \autoref{fig:cuda-op}. We include TensorComprehension (TC, commit: ef644ba)~\cite{TC}, a recently introduced auto-tuning framework, as an additional baseline.~\footnote{According to personal communication~\cite{TCcomment}, TC is not yet meant to be used for compute-bound problems. However, it is still a good reference baseline to include in the comparison.}
TC results include the best kernels it found in $10$ generations $\times$ $100$ population $\times$ $2$ random seeds for each operator  (i.e., 2000 trials per operator).
2D convolution, one of the most important DL operators, is heavily optimized by cuDNN.
However, \TensorOpt can still generate better GPU kernels for most layers. Depthwise convolution is a newly introduced operator with a simpler structure~\cite{Howard:MobileNet}. In this case, both \TensorOpt and TC can find fast kernels compared to MXNet's handcrafted kernels.
\TensorOpt's improvements are mainly due to its exploration of a large schedule space and an effective ML-based search algorithm.

\subsection{Embedded CPU Evaluation}

\begin{figure}[t]
	\centering
	\includegraphics[width=1\columnwidth]{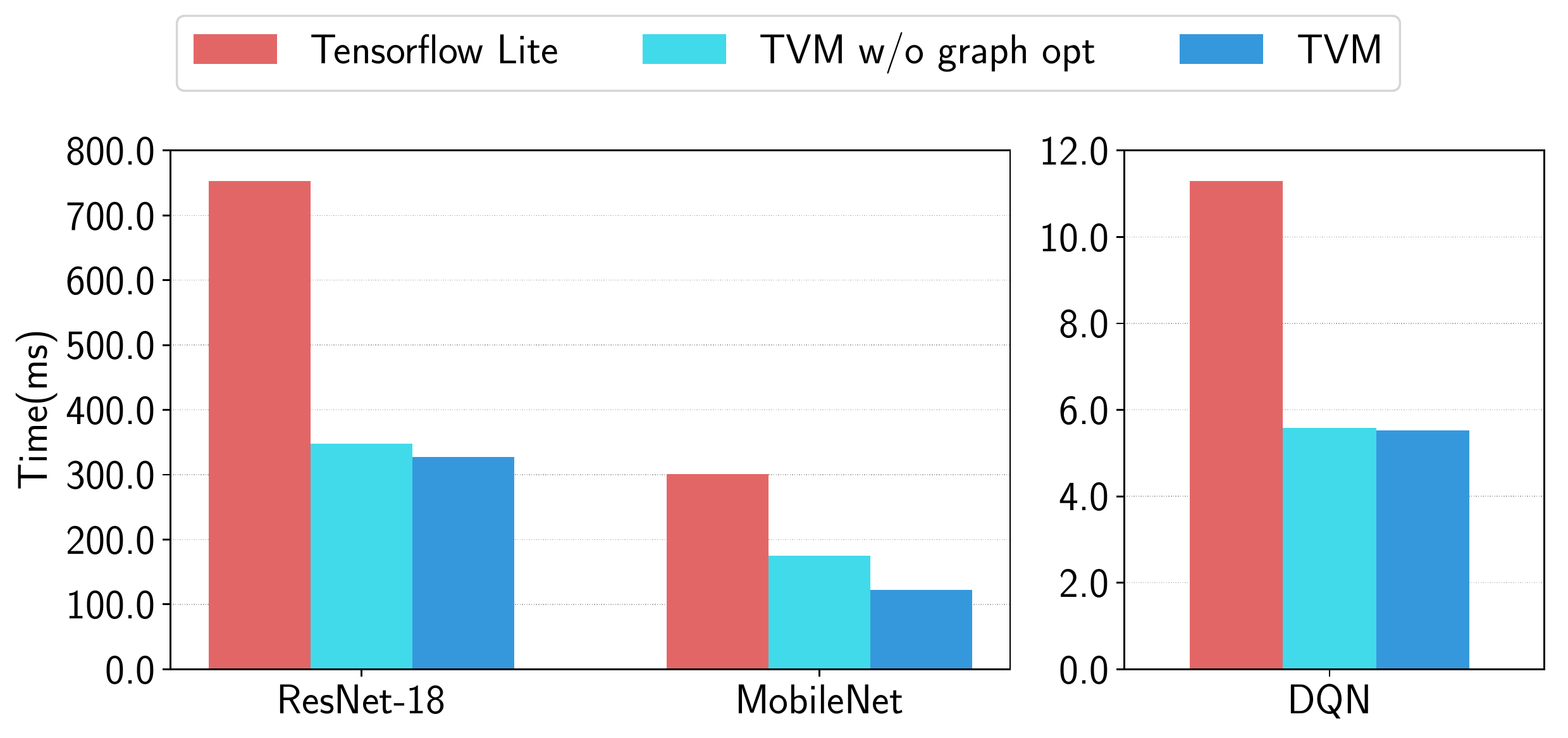}
	\precap
	\caption{\small{ARM A53 end-to-end evaluation of \TensorOpt and TFLite.}}
	\postcap
	\label{fig:rasp-e2e}
\end{figure}

\begin{figure}[t]
	\centering
	\includegraphics[width=\columnwidth]{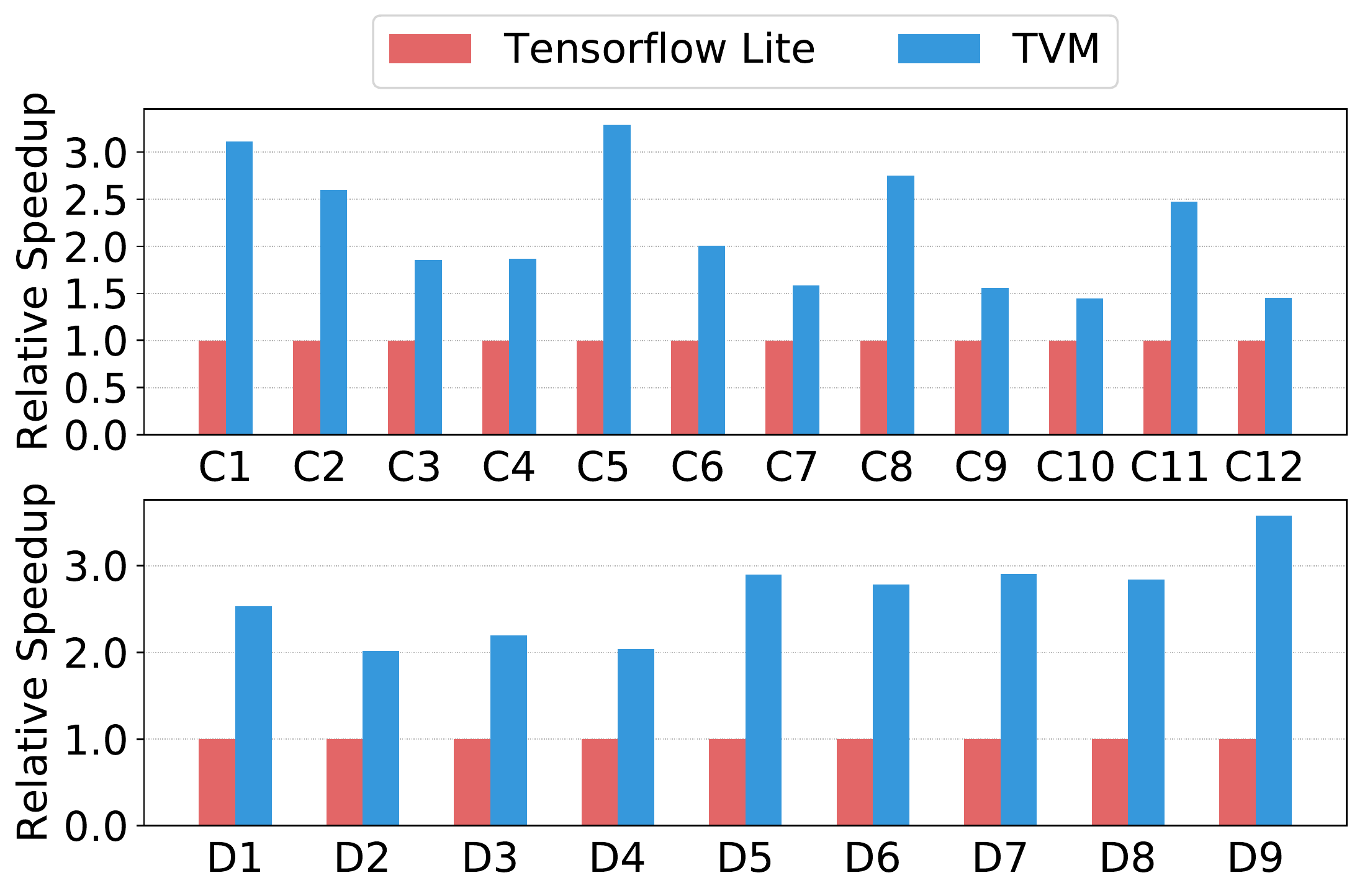}
	\precap
	\caption{\capsize{Relative speedup of all conv2d operators in ResNet-18 and all depthwise conv2d operators in mobilenet. Tested on ARM A53. See \autoref{tbl:all-op}  for the configurations of these operators.}}
	\postcap
	\label{fig:rasp-op}
\end{figure}

We evaluated the performance of \TensorOpt on an ARM Cortex A53 (Quad Core 1.2GHz).
We used Tensorflow Lite (TFLite, commit: 7558b085) as our baseline system.
\autoref{fig:rasp-op} compares \TensorOpt operators
to hand-optimized ones for ResNet and MobileNet.
We observe that \TensorOpt generates operators that outperform the hand-optimized TFLite versions for both neural network workloads.
This result also demonstrates \TensorOpt's ability to quickly optimize emerging
tensor operators, such as depthwise convolution operators.
Finally, ~\autoref{fig:rasp-e2e} shows an end-to-end comparison of three workloads, where \TensorOpt outperforms the TFLite baseline.\footnote{DCGAN and LSTM results are not presented because they are not yet supported by the baseline.}

%Expressing fine-grained data packing requirements is simplified with \TensorOpt: by treating the bits that compose a word as an additional dimension, we can express complex data packing operations as simple tensor dimension re-ordering operations.

\paragraph{Ultra Low-Precision Operators} We demonstrate \TensorOpt's ability to support ultra low-precision inference~\cite{Courbariaux:BinaryConnect, rastegari2016xnor} by generating highly optimized operators for fixed-point data types of less than 8-bits.
Low-precision networks replace expensive multiplication with vectorized bit-serial multiplication that is composed of bitwise \textit{and} popcount reductions~\cite{tulloch2017high}.
Achieving efficient low-precision inference requires packing quantized data types into wider standard data types, such as \texttt{int8} or \texttt{int32}.
Our system generates code that outperforms hand-optimized libraries from Caffe2 (commit: 39e07f7)\cite{tulloch2017high}.
We implemented an ARM-specific \emph{tensorization} intrinsic that leverages ARM instructions to build an efficient, low-precision matrix-vector microkernel.% in just \TODO{lines of code}.
 We then used \TensorOpt's automated optimizer to explore the scheduling space.

\begin{figure}[t]
	\centering
	\includegraphics[width=.9\columnwidth]{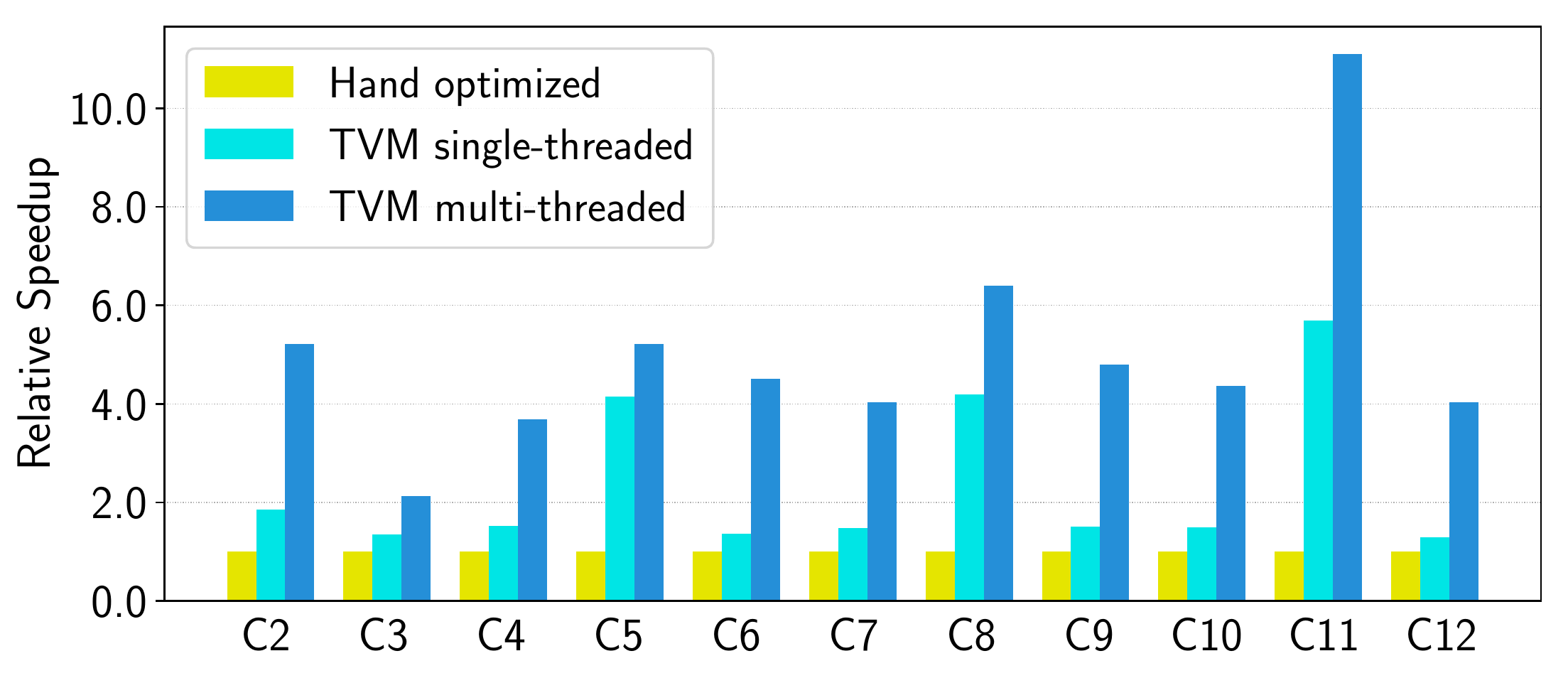}
	\precap
	\caption{\capsize{Relative speedup of single- and multi-threaded low-precision conv2d operators in ResNet. Baseline was a single-threaded, hand-optimized implementation from Caffe2~(commit: 39e07f7).
\REV{C5, C3 are 1x1 convolutions that have less compute intensity, resulting in less speedup by multi-threading.}
}}
	\postcap
	\label{fig:rasp-qnn}
\end{figure}

Figure \ref{fig:rasp-qnn} compares \TensorOpt to the Caffe2 ultra low-precision library on ResNet for 2-bit activations, 1-bit weights inference.
Since the baseline is single threaded, we also compare it to a single-threaded \TensorOpt version.
% demonstrating \TensorOpt's ability to match the performance of hand optimized libraries.
Single-threaded \TensorOpt outperforms the baseline, particularly for \texttt{C5}, \texttt{C8}, and \texttt{C11} layers; these are convolution layers of kernel size $1\times1$ and stride of 2 for which  the ultra low-precision baseline library is not optimized.
Furthermore, we take advantage of additional \TensorOpt capabilities to produce a parallel library implementation that shows improvement over the baseline.
In addition to the 2-bit+1-bit configuration, \TensorOpt can generate and optimize for other precision configurations that are unsupported by the baseline library, offering improved flexibility.

\subsection{Embedded GPU Evaluation}
\begin{figure}[t]
 \centering
 \includegraphics[width=\columnwidth]{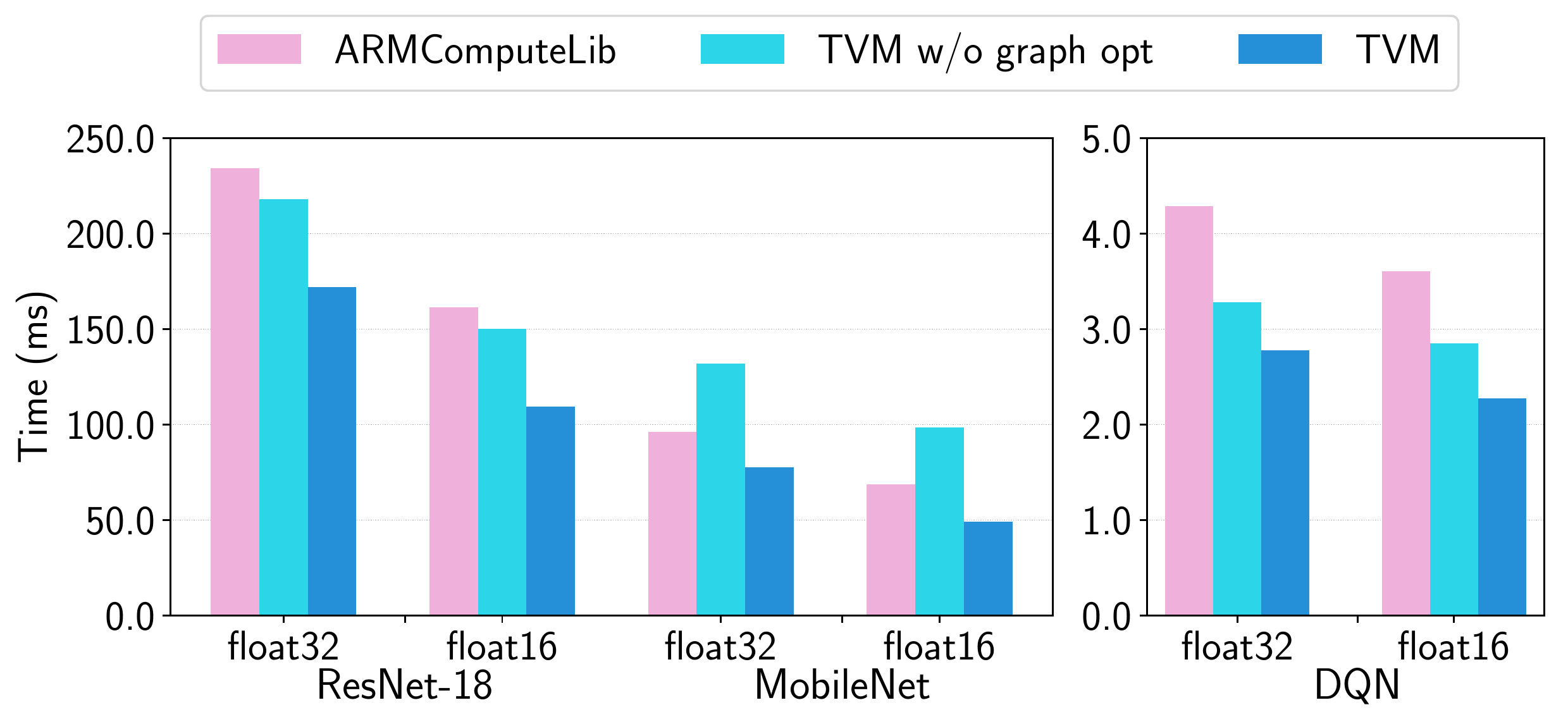}
 \precap
 \caption{\capsize{End-to-end experiment results on Mali-T860MP4. Two data types, float32 and float16, were evaluated.}}
 \postcap
 \label{fig:mali-end2end}
\end{figure}

For our mobile GPU experiments, we ran our end-to-end pipeline on a Firefly-RK3399 board equipped with an ARM Mali-T860MP4 GPU. The baseline was a vendor-provided library, the ARM Compute Library (v18.03).
As shown in \autoref{fig:mali-end2end}, we outperformed the baseline on three available models for both \texttt{float16} and \texttt{float32} (DCGAN and LSTM are not yet supported by the baseline). The speedup ranged from 1.2$\times$ to 1.6$\times$.

\subsection{FPGA Accelerator Evaluation}
\label{sec:fpga_eval}

\begin{figure}[t]
\centering
\includegraphics[width=\columnwidth]{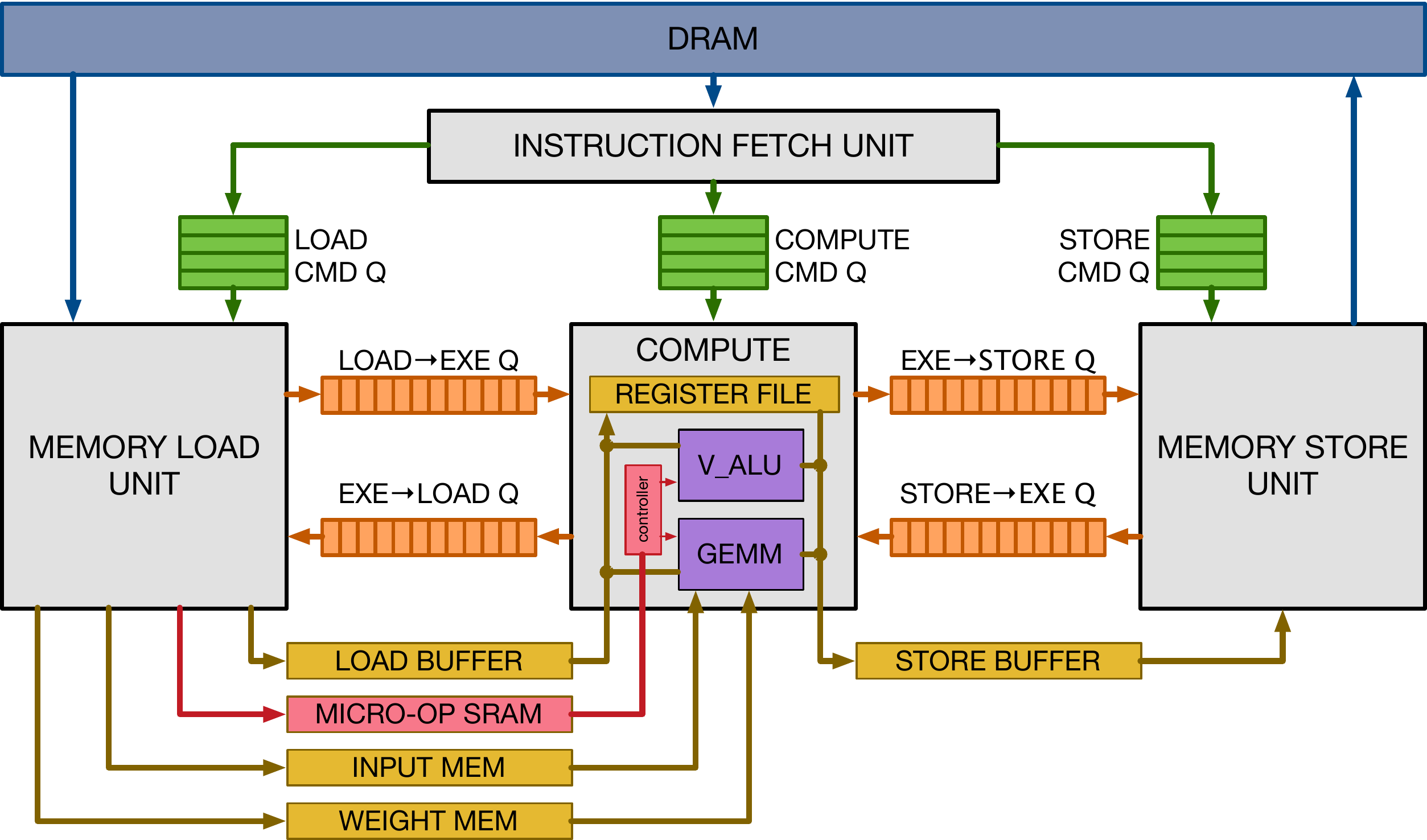}
\precap
\caption{\accel Hardware design overview.}
\postcap
\label{fig:vdla_overview}
\end{figure}

\paragraph{\accelmeaning}
We now relate how \TensorOpt tackled accelerator-specific code generation on a generic inference accelerator design we prototyped on an FPGA.
We used in this evaluation the \accelmeaning~(\accelno) -- which distills characteristics from previous accelerator proposals~\cite{Liu:PuDianNao, Chen:Eyeriss, Jouppi:TPU} into a minimalist hardware architecture --   to demonstrate \TensorOpt's ability to generate highly efficient schedules that can target specialized accelerators.
~\autoref{fig:vdla_overview} shows the high-level hardware organization of the \accel architecture.
\accel is programmed
%as a vector processor on  operators with low compute intensity (activation, batch norm), and
as a tensor processor to efficiently execute operations with high compute intensity (e.g, matrix multiplication, high dimensional convolution).
It can perform load/store operations to bring blocked 3-dimensional tensors from DRAM into a contiguous region of SRAM.
It also provides specialized on-chip memories for network parameters, layer inputs (narrow data type), and layer outputs (wide data type).
Finally, \accel provides explicit synchronization control over successive loads, computes, and stores to maximize the overlap between memory and compute operations.

\paragraph{Methodology.}
We implemented the \accel design on a low-power PYNQ board that incorporates an ARM Cortex A9 dual core CPU clocked at 667MHz and an Artix-7 based FPGA fabric.
On these modest FPGA resources, we implemented a $16\times16$ matrix-vector unit clocked at 200MHz that performs products of 8-bit values and accumulates them into a 32-bit register every cycle.
The theoretical peak throughput of this \accel design is about 102.4GOPS/s.
We allocated 32kB of resources for activation storage, 32kB for parameter storage, 32kB for microcode buffers, and 128kB for the register file.
These on-chip buffers are by no means large enough to provide sufficient on-chip storage for a single layer of ResNet and therefore enable a case study on effective memory reuse and latency hiding.

We built a driver library for \accel with a C runtime API that
constructs instructions and pushes them to the target accelerator for execution.
Our code generation algorithm then translates the accelerator program to
a series of calls into the runtime API.  Adding the specialized accelerator back-end took $\sim$2k LoC in Python.

\paragraph{End-to-End ResNet Evaluation.}
We used \TensorOpt to generate ResNet inference kernels on the PYNQ platform and offloaded as many layers as possible to \accelno.
We also used it to generate both schedules for the CPU only and CPU+FPGA implementation.
Due to its shallow convolution depth, the first ResNet convolution layer could not be efficiently offloaded on the FPGA and was instead computed on the CPU.
All other convolution layers in ResNet, however, were amenable to efficient offloading.
Operations like residual layers and activations were also performed on the CPU since \accel does not support these operations.

\begin{figure}[t]
\centering
\includegraphics[width=1\columnwidth]{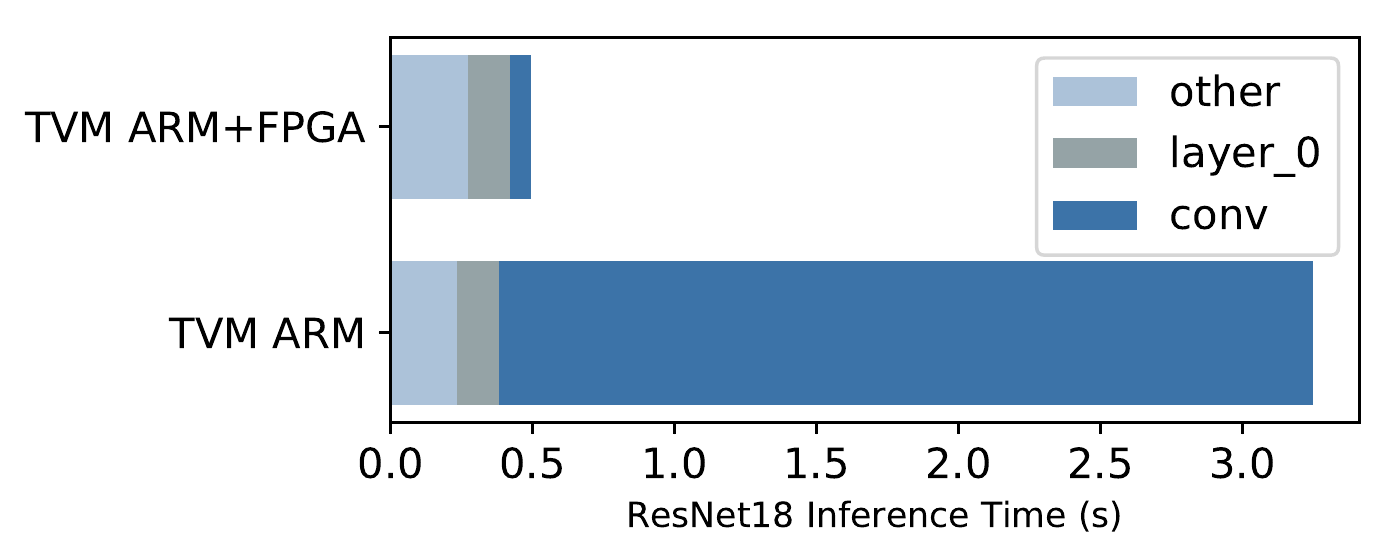}
\precap
\caption{\capsize{We offloaded convolutions in the ResNet workload to an FPGA-based accelerator.
The grayed-out bars correspond to layers that could not be accelerated by the FPGA and therefore had to run on the CPU. The FPGA provided a 40x acceleration on offloaded convolution layers over the Cortex A9.}}
\postcap
\label{fig:e2e_fpga}
\end{figure}

~\autoref{fig:e2e_fpga} breaks down ResNet inference time into CPU-only execution and CPU+FPGA execution.
Most computation was spent on the convolution layers that could be offloaded to \accelno.
For those convolution layers, the achieved speedup was 40$\times$.
Unfortunately, due to Amdahl's law, the overall performance of the FPGA accelerated system was bottlenecked by the sections of the workload that had to be executed on the CPU.
We envision that extending the \accel design to support these other operators will help reduce cost even further.
This FPGA-based experiment showcases \TensorOptno's ability to adapt to new architectures and the hardware intrinsics they expose.

% \begin{table}
% \begin{tabular}{lllllrrr}
% \hline
% H/W & IC & OC & K & S & \parbox[r]{.14\linewidth}{ARM Cortex A9 CPU\\(ms)} & \parbox[r]{.14\linewidth}{FPGA with Latency Hiding\\(ms)} & \parbox[r]{.13\linewidth}{FPGA w/o Latency Hiding\\(ms)} & Speedup \\ \hline
% 224 & 3 & 64 & 7 & 2 & 151 & n/a & n/a & n/a \\
% 56 & 64 & 64 & 3 & 1 & 197 & 3.77 & 4.73 & 1.25 \\
% 56 & 64 & 64 & 1 & 1 & 17.4 & 1.53 & 1.71 & 1.12 \\
% 56 & 64 & 128 & 3 & 2 & 70.7 & 1.90 & 2.71 & 1.43 \\
% 56 & 64 & 128 & 1 & 2 & 13.1 & 1.19 & 1.40 & 1.17 \\
% 28 & 128 & 128 & 3 & 1 & 197 & 3.04 & 4.09 & 1.35 \\
% 28 & 128 & 256 & 3 & 2 & 64.3 & 2.12 & 2.38 & 1.12 \\
% 28 & 128 & 256 & 1 & 2 & 13.2 & 0.69 & 0.83 & 1.20 \\
% 14 & 256 & 256 & 3 & 1 & 197 & 3.49 & 5.53 & 1.58 \\
% 14 & 256 & 512 & 3 & 2 & 107 & 4.43 & 5.08 & 1.15 \\
% 14 & 256 & 512 & 1 & 2 & 9.67 & 0.83 & 0.92 & 1.10 \\
% 7 & 512 & 512 & 3 & 1 & 216 & n/s & 9.51 & n/a \\
% \hline
% \end{tabular}
% \caption{\TensorOpt optimizing ResNet for explicit latency hiding on the FPGA-based PYNQ. H/W for height and width, IC for input channels, OC for output channels, K for kernel size, and S for stride size.}
% \label{tab:fpga-res}
% \end{table}

\section{Related Work}
\label{sec:rel}

Deep learning frameworks~\cite{tensorflow-osdi,Bastien-Theano-2012,MXNet-whitepaper,CNTK}
provide convenient interfaces for users to express DL workloads and deploy them easily on different hardware back-ends.
While existing frameworks currently depend on vendor-specific tensor operator libraries to execute their workloads, they can leverage \TensorOpt's stack to generate optimized code for a larger number of hardware devices.

High-level computation graph DSLs are a typical way to represent and perform high-level optimizations.
Tensorflow's XLA~\cite{tensorflow-osdi} and the recently introduced DLVM~\cite{Wei:DLVM} fall into this category.
The representations of computation graphs in these works are similar, and a high-level computation graph DSL is also used in this paper.
While graph-level representations are a good fit for high-level optimizations,
they are too high level to optimize tensor operators under a diverse set of hardware back-ends.
Prior work relies on specific lowering rules to directly
generate low-level LLVM or resorts to vendor-crafted libraries.
These approaches require significant engineering effort for each hardware back-end and operator-variant combination.

Halide~\cite{JRK:Halide} introduced the idea of separating computing and scheduling.
We adopt Halide's insights and reuse its existing useful scheduling primitives in our compiler.
Our tensor operator scheduling is also related to other work on DSL for GPUs~\cite{Steuwer:Lift,Henriksen:Futhark,OptiML,Loopy}
and polyhedral-based loop transformation~\cite{Baghdadi:PENCIL,Verdoolaege:PPCG}.
TACO~\cite{Kjolstad:TACO} introduces a generic way to generate sparse tensor operators on CPU.
Weld~\cite{Weld} is a DSL for data processing tasks.
We specifically focus on solving the new scheduling challenges of DL workloads for GPUs and specialized accelerators.
Our new primitives can potentially be adopted by the optimization pipelines in these works.

High-performance libraries such as ATLAS~\cite{Whaley:ATLAS} and
FFTW~\cite{FFTW} use auto-tuning to get the best performance.
Tensor comprehension~\cite{TC} applied black-box auto-tuning together with polyhedral optimizations to optimize CUDA kernels.
\REV{OpenTuner~\cite{OpenTuner} and existing hyper parameter-tuning algorithms~\cite{Li:Hyperband} apply domain-agnostic search.}
A predefined cost model is used to automatically schedule image processing pipelines
in Halide~\cite{Mullapudi:AutomaticScheduling}.
\REV{\TensorOpt's ML model uses effective domain-aware cost modeling that considers program structure.}
The based distributed schedule optimizer scales to a larger search space and
can find state-of-the-art kernels on a large range of supported back-ends.
More importantly, we provide an end-to-end stack that can take descriptions directly from DL frameworks
and jointly optimize together with the graph-level stack.

Despite the emerging popularity of accelerators for deep learning~\cite{Jouppi:TPU,Chen:DaDianNao},
it remains unclear how a compilation stack can be built to effectively target these devices.
The \accel design used in our evaluation provides a generic way to summarize the properties of TPU-like accelerators
and enables a concrete case study on how to compile code for accelerators.
\REV{Our approach could potentially benefit existing systems that
  compile deep learning to FPGA~\cite{dnnweaver:micro16,Umuroglu:FINN}, as well.}
This paper provides a generic solution to effectively target accelerators via tensorization and compiler-driven latency hiding.

\section{Conclusion}
\label{sec:con}
We proposed an end-to-end compilation stack to solve fundamental optimization challenges for deep learning across a diverse set of hardware back-ends.  Our system includes automated end-to-end optimization, which is historically a labor-intensive and highly specialized task.
We hope this  work will encourage additional studies of end-to-end compilation approaches and open new opportunities for DL system software-hardware co-design techniques.

\section*{Acknowledgement}

We would like to thank Ras Bodik, James Bornholt, Xi Wang, Tom Anderson and Qiao Zhang
for their thorough feedback on earlier versions of this paper. We would also like to thank members of Sampa, SAMPL and Systems groups at the Allen School for their feedback on the work and manuscript.
We would like to thank the anonymous OSDI reviewers, and our shepherd, Ranjita Bhagwan, for helpful feedbacks. This work was supported in part by a Google PhD Fellowship for Tianqi Chen,
ONR award \#N00014-16-1-2795, NSF under grants  CCF-1518703, CNS-1614717, and CCF-1723352, and gifts from Intel (under the CAPA program), Oracle, Huawei and anonymous sources.

{\footnotesize \bibliographystyle{acm}
\bibliography{TensorOpt}}

\begin{thebibliography}{10}

\bibitem{volta-whitepaper}
{NVIDIA Tesla V100 GPU Architecture: The World's Most Advanced Data Center
  GPU}, 2017.

\bibitem{tensorflow2015-whitepaper}
{\sc Abadi, M., Agarwal, A., Barham, P., Brevdo, E., Chen, Z., Citro, C.,
  Corrado, G.~S., Davis, A., Dean, J., Devin, M., Ghemawat, S., Goodfellow, I.,
  Harp, A., Irving, G., Isard, M., Jia, Y., Jozefowicz, R., Kaiser, L., Kudlur,
  M., Levenberg, J., Man\'{e}, D., Monga, R., Moore, S., Murray, D., Olah, C.,
  Schuster, M., Shlens, J., Steiner, B., Sutskever, I., Talwar, K., Tucker, P.,
  Vanhoucke, V., Vasudevan, V., Vi\'{e}gas, F., Vinyals, O., Warden, P.,
  Wattenberg, M., Wicke, M., Yu, Y., and Zheng, X.}
\newblock {TensorFlow}: Large-scale machine learning on heterogeneous systems,
  2015.
\newblock Software available from tensorflow.org.

\bibitem{tensorflow-osdi}
{\sc Abadi, M., Barham, P., Chen, J., Chen, Z., Davis, A., Dean, J., Devin, M.,
  Ghemawat, S., Irving, G., Isard, M., Kudlur, M., Levenberg, J., Monga, R.,
  Moore, S., Murray, D.~G., Steiner, B., Tucker, P., Vasudevan, V., Warden, P.,
  Wicke, M., Yu, Y., and Zheng, X.}
\newblock Tensorflow: A system for large-scale machine learning.
\newblock In {\em 12th USENIX Symposium on Operating Systems Design and
  Implementation (OSDI 16)\/} (2016), pp.~265--283.

\bibitem{CNTK}
{\sc Agarwal, A., Akchurin, E., Basoglu, C., Chen, G., Cyphers, S., Droppo, J.,
  Eversole, A., Guenter, B., Hillebrand, M., Hoens, R., Huang, X., Huang, Z.,
  Ivanov, V., Kamenev, A., Kranen, P., Kuchaiev, O., Manousek, W., May, A.,
  Mitra, B., Nano, O., Navarro, G., Orlov, A., Padmilac, M., Parthasarathi, H.,
  Peng, B., Reznichenko, A., Seide, F., Seltzer, M.~L., Slaney, M., Stolcke,
  A., Wang, Y., Wang, H., Yao, K., Yu, D., Zhang, Y., and Zweig, G.}
\newblock An introduction to computational networks and the computational
  network toolkit.
\newblock Tech. Rep. MSR-TR-2014-112, August 2014.

\bibitem{OpenTuner}
{\sc Ansel, J., Kamil, S., Veeramachaneni, K., Ragan-Kelley, J., Bosboom, J.,
  O'Reilly, U.-M., and Amarasinghe, S.}
\newblock Opentuner: An extensible framework for program autotuning.
\newblock In {\em International Conference on Parallel Architectures and
  Compilation Techniques\/} (Edmonton, Canada, August 2014).

\bibitem{Baghdadi:PENCIL}
{\sc Baghdadi, R., Beaugnon, U., Cohen, A., Grosser, T., Kruse, M., Reddy, C.,
  Verdoolaege, S., Betts, A., Donaldson, A.~F., Ketema, J., Absar, J.,
  Haastregt, S.~v., Kravets, A., Lokhmotov, A., David, R., and Hajiyev, E.}
\newblock Pencil: A platform-neutral compute intermediate language for
  accelerator programming.
\newblock In {\em Proceedings of the 2015 International Conference on Parallel
  Architecture and Compilation (PACT)\/} (Washington, DC, USA, 2015), PACT '15,
  IEEE Computer Society, pp.~138--149.

\bibitem{Bastien-Theano-2012}
{\sc Bastien, F., Lamblin, P., Pascanu, R., Bergstra, J., Goodfellow, I.~J.,
  Bergeron, A., Bouchard, N., and Bengio, Y.}
\newblock Theano: new features and speed improvements.
\newblock Deep Learning and Unsupervised Feature Learning NIPS 2012 Workshop,
  2012.

\bibitem{XGBoostKDD}
{\sc Chen, T., and Guestrin, C.}
\newblock Xgboost: A scalable tree boosting system.
\newblock In {\em Proceedings of the 22Nd ACM SIGKDD International Conference
  on Knowledge Discovery and Data Mining\/} (New York, NY, USA, 2016), KDD '16,
  ACM, pp.~785--794.

\bibitem{MXNet-whitepaper}
{\sc Chen, T., Li, M., Li, Y., Lin, M., Wang, N., Wang, M., Xiao, T., Xu, B.,
  Zhang, C., , and Zhang, Z.}
\newblock {MXNet}: A flexible and efficient machine learning library for
  heterogeneous distributed systems.
\newblock In {\em Neural Information Processing Systems, Workshop on Machine
  Learning Systems (LearningSys'15)\/} (2015).

\bibitem{Chen:Prefetching}
{\sc Chen, T.-F., and Baer, J.-L.}
\newblock Effective hardware-based data prefetching for high-performance
  processors.
\newblock {\em IEEE Transactions on Computers 44}, 5 (May 1995), 609--623.

\bibitem{Chen:DaDianNao}
{\sc Chen, Y., Luo, T., Liu, S., Zhang, S., He, L., Wang, J., Li, L., Chen, T.,
  Xu, Z., Sun, N., and Temam, O.}
\newblock Dadiannao: A machine-learning supercomputer.
\newblock In {\em Proceedings of the 47th Annual IEEE/ACM International
  Symposium on Microarchitecture\/} (Washington, DC, USA, 2014), MICRO-47, IEEE
  Computer Society, pp.~609--622.

\bibitem{Chen:Eyeriss}
{\sc Chen, Y.-H., Emer, J., and Sze, V.}
\newblock Eyeriss: A spatial architecture for energy-efficient dataflow for
  convolutional neural networks.
\newblock In {\em Proceedings of the 43rd International Symposium on Computer
  Architecture\/} (Piscataway, NJ, USA, 2016), ISCA '16, IEEE Press,
  pp.~367--379.

\bibitem{Courbariaux:BinaryConnect}
{\sc Courbariaux, M., Bengio, Y., and David, J.}
\newblock Binaryconnect: Training deep neural networks with binary weights
  during propagations.
\newblock {\em CoRR abs/1511.00363\/} (2015).

\bibitem{Eggers:SMT}
{\sc Eggers, S.~J., Emer, J.~S., Levy, H.~M., Lo, J.~L., Stamm, R.~L., and
  Tullsen, D.~M.}
\newblock Simultaneous multithreading: a platform for next-generation
  processors.
\newblock {\em IEEE Micro 17}, 5 (Sept 1997), 12--19.

\bibitem{FFTW}
{\sc Frigo, M., and Johnson, S.~G.}
\newblock Fftw: an adaptive software architecture for the fft.
\newblock In {\em Acoustics, Speech and Signal Processing, 1998. Proceedings of
  the 1998 IEEE International Conference on\/} (May 1998), vol.~3,
  pp.~1381--1384 vol.3.

\bibitem{He2016}
{\sc He, K., Zhang, X., Ren, S., and Sun, J.}
\newblock Identity mappings in deep residual networks.
\newblock {\em arXiv preprint arXiv:1603.05027\/} (2016).

\bibitem{Hegarty:DarkRoom}
{\sc Hegarty, J., Brunhaver, J., DeVito, Z., Ragan-Kelley, J., Cohen, N., Bell,
  S., Vasilyev, A., Horowitz, M., and Hanrahan, P.}
\newblock Darkroom: Compiling high-level image processing code into hardware
  pipelines.
\newblock {\em ACM Trans. Graph. 33}, 4 (July 2014), 144:1--144:11.

\bibitem{Henriksen:Futhark}
{\sc Henriksen, T., Serup, N. G.~W., Elsman, M., Henglein, F., and Oancea,
  C.~E.}
\newblock Futhark: Purely functional gpu-programming with nested parallelism
  and in-place array updates.
\newblock In {\em Proceedings of the 38th ACM SIGPLAN Conference on Programming
  Language Design and Implementation\/} (New York, NY, USA, 2017), PLDI 2017,
  ACM, pp.~556--571.

\bibitem{Howard:MobileNet}
{\sc Howard, A.~G., Zhu, M., Chen, B., Kalenichenko, D., Wang, W., Weyand, T.,
  Andreetto, M., and Adam, H.}
\newblock Mobilenets: Efficient convolutional neural networks for mobile vision
  applications.
\newblock {\em CoRR abs/1704.04861\/} (2017).

\bibitem{Jouppi:CachePerf}
{\sc Jouppi, N.~P.}
\newblock Improving direct-mapped cache performance by the addition of a small
  fully-associative cache and prefetch buffers.
\newblock In {\em [1990] Proceedings. The 17th Annual International Symposium
  on Computer Architecture\/} (May 1990), pp.~364--373.

\bibitem{Jouppi:TPU}
{\sc Jouppi, N.~P., Young, C., Patil, N., Patterson, D., Agrawal, G., Bajwa,
  R., Bates, S., Bhatia, S., Boden, N., Borchers, A., Boyle, R., Cantin, P.-l.,
  Chao, C., Clark, C., Coriell, J., Daley, M., Dau, M., Dean, J., Gelb, B.,
  Ghaemmaghami, T.~V., Gottipati, R., Gulland, W., Hagmann, R., Ho, C.~R.,
  Hogberg, D., Hu, J., Hundt, R., Hurt, D., Ibarz, J., Jaffey, A., Jaworski,
  A., Kaplan, A., Khaitan, H., Killebrew, D., Koch, A., Kumar, N., Lacy, S.,
  Laudon, J., Law, J., Le, D., Leary, C., Liu, Z., Lucke, K., Lundin, A.,
  MacKean, G., Maggiore, A., Mahony, M., Miller, K., Nagarajan, R.,
  Narayanaswami, R., Ni, R., Nix, K., Norrie, T., Omernick, M., Penukonda, N.,
  Phelps, A., Ross, J., Ross, M., Salek, A., Samadiani, E., Severn, C.,
  Sizikov, G., Snelham, M., Souter, J., Steinberg, D., Swing, A., Tan, M.,
  Thorson, G., Tian, B., Toma, H., Tuttle, E., Vasudevan, V., Walter, R., Wang,
  W., Wilcox, E., and Yoon, D.~H.}
\newblock In-datacenter performance analysis of a tensor processing unit.
\newblock In {\em Proceedings of the 44th Annual International Symposium on
  Computer Architecture\/} (New York, NY, USA, 2017), ISCA '17, ACM, pp.~1--12.

\bibitem{SimulatedAnealing}
{\sc Kirkpatrick, S., Gelatt, C.~D., and Vecchi, M.~P.}
\newblock Optimization by simulated annealing.
\newblock {\em Science 220}, 4598 (1983), 671--680.

\bibitem{Kjolstad:TACO}
{\sc Kjolstad, F., Kamil, S., Chou, S., Lugato, D., and Amarasinghe, S.}
\newblock The tensor algebra compiler.
\newblock {\em Proc. ACM Program. Lang. 1}, OOPSLA (Oct. 2017), 77:1--77:29.

\bibitem{Loopy}
{\sc {Kl{\"o}ckner}, A.}
\newblock {Loo.py: transformation-based code~generation for GPUs and CPUs}.
\newblock In {\em {Proceedings of ARRAY `14: ACM SIGPLAN Workshop on Libraries,
  Languages, and Compilers for Array Programming}\/} ({Edinburgh, Scotland.},
  2014), {Association for Computing Machinery}.

\bibitem{Winograd}
{\sc Lavin, A., and Gray, S.}
\newblock Fast algorithms for convolutional neural networks.
\newblock In {\em 2016 {IEEE} Conference on Computer Vision and Pattern
  Recognition, {CVPR} 2016, Las Vegas, NV, USA, June 27-30, 2016\/} (2016),
  pp.~4013--4021.

\bibitem{Li:Hyperband}
{\sc Li, L., Jamieson, K.~G., DeSalvo, G., Rostamizadeh, A., and Talwalkar, A.}
\newblock Efficient hyperparameter optimization and infinitely many armed
  bandits.
\newblock {\em CoRR abs/1603.06560\/} (2016).

\bibitem{Liu:PuDianNao}
{\sc Liu, D., Chen, T., Liu, S., Zhou, J., Zhou, S., Teman, O., Feng, X., Zhou,
  X., and Chen, Y.}
\newblock Pudiannao: A polyvalent machine learning accelerator.
\newblock In {\em Proceedings of the Twentieth International Conference on
  Architectural Support for Programming Languages and Operating Systems\/} (New
  York, NY, USA, 2015), ASPLOS '15, ACM, pp.~369--381.

\bibitem{mnih2015human}
{\sc Mnih, V., Kavukcuoglu, K., Silver, D., Rusu, A.~A., Veness, J., Bellemare,
  M.~G., Graves, A., Riedmiller, M., Fidjeland, A.~K., Ostrovski, G., et~al.}
\newblock Human-level control through deep reinforcement learning.
\newblock {\em Nature 518}, 7540 (2015), 529.

\bibitem{Mullapudi:AutomaticScheduling}
{\sc Mullapudi, R.~T., Adams, A., Sharlet, D., Ragan-Kelley, J., and
  Fatahalian, K.}
\newblock Automatically scheduling halide image processing pipelines.
\newblock {\em ACM Trans. Graph. 35}, 4 (July 2016), 83:1--83:11.

\bibitem{Weld}
{\sc Palkar, S., Thomas, J.~J., Narayanan, D., Shanbhag, A., Palamuttam, R.,
  Pirk, H., Schwarzkopf, M., Amarasinghe, S.~P., Madden, S., and Zaharia, M.}
\newblock Weld: Rethinking the interface between data-intensive applications.
\newblock {\em CoRR abs/1709.06416\/} (2017).

\bibitem{radford2015dcgan}
{\sc Radford, A., Metz, L., and Chintala, S.}
\newblock Unsupervised representation learning with deep convolutional
  generative adversarial networks.
\newblock {\em arXiv preprint arXiv:1511.06434\/} (2015).

\bibitem{JRK:Halide}
{\sc Ragan-Kelley, J., Barnes, C., Adams, A., Paris, S., Durand, F., and
  Amarasinghe, S.}
\newblock Halide: A language and compiler for optimizing parallelism, locality,
  and recomputation in image processing pipelines.
\newblock In {\em Proceedings of the 34th ACM SIGPLAN Conference on Programming
  Language Design and Implementation\/} (New York, NY, USA, 2013), PLDI '13,
  ACM, pp.~519--530.

\bibitem{rastegari2016xnor}
{\sc Rastegari, M., Ordonez, V., Redmon, J., and Farhadi, A.}
\newblock Xnor-net: Imagenet classification using binary convolutional neural
  networks.
\newblock In {\em European Conference on Computer Vision\/} (2016), Springer,
  pp.~525--542.

\bibitem{dnnweaver:micro16}
{\sc Sharma, H., Park, J., Mahajan, D., Amaro, E., Kim, J.~K., Shao, C.,
  Mishra, A., and Esmaeilzadeh, H.}
\newblock From high-level deep neural models to fpgas.
\newblock In {\em Microarchitecture (MICRO), 2016 49th Annual IEEE/ACM
  International Symposium on\/} (2016), IEEE, pp.~1--12.

\bibitem{Smith:DAE}
{\sc Smith, J.~E.}
\newblock Decoupled access/execute computer architectures.
\newblock In {\em Proceedings of the 9th Annual Symposium on Computer
  Architecture\/} (Los Alamitos, CA, USA, 1982), ISCA '82, IEEE Computer
  Society Press, pp.~112--119.

\bibitem{Steuwer:Lift}
{\sc Steuwer, M., Remmelg, T., and Dubach, C.}
\newblock Lift: A functional data-parallel ir for high-performance gpu code
  generation.
\newblock In {\em Proceedings of the 2017 International Symposium on Code
  Generation and Optimization\/} (Piscataway, NJ, USA, 2017), CGO '17, IEEE
  Press, pp.~74--85.

\bibitem{OptiML}
{\sc Sujeeth, A.~K., Lee, H., Brown, K.~J., Chafi, H., Wu, M., Atreya, A.~R.,
  Olukotun, K., Rompf, T., and Odersky, M.}
\newblock Optiml: An implicitly parallel domain-specific language for machine
  learning.
\newblock In {\em Proceedings of the 28th International Conference on
  International Conference on Machine Learning\/} (USA, 2011), ICML'11,
  pp.~609--616.

\bibitem{TreeRNN}
{\sc Tai, K.~S., Socher, R., and Manning, C.~D.}
\newblock Improved semantic representations from tree-structured long
  short-term memory networks.
\newblock {\em arXiv preprint arXiv:1503.00075\/} (2015).

\bibitem{tulloch2017high}
{\sc Tulloch, A., and Jia, Y.}
\newblock High performance ultra-low-precision convolutions on mobile devices.
\newblock {\em arXiv preprint arXiv:1712.02427\/} (2017).

\bibitem{Umuroglu:FINN}
{\sc Umuroglu, Y., Fraser, N.~J., Gambardella, G., Blott, M., Leong, P. H.~W.,
  Jahre, M., and Vissers, K.~A.}
\newblock {FINN:} {A} framework for fast, scalable binarized neural network
  inference.
\newblock {\em CoRR abs/1612.07119\/} (2016).

\bibitem{TCcomment}
{\sc Vasilache, N.}
\newblock personal communication.

\bibitem{TC}
{\sc Vasilache, N., Zinenko, O., Theodoridis, T., Goyal, P., DeVito, Z., Moses,
  W.~S., Verdoolaege, S., Adams, A., and Cohen, A.}
\newblock Tensor comprehensions: Framework-agnostic high-performance machine
  learning abstractions.
\newblock {\em CoRR abs/1802.04730\/} (2018).

\bibitem{Verdoolaege:PPCG}
{\sc Verdoolaege, S., Carlos~Juega, J., Cohen, A., Ignacio~G\'{o}mez, J.,
  Tenllado, C., and Catthoor, F.}
\newblock Polyhedral parallel code generation for cuda.
\newblock {\em ACM Trans. Archit. Code Optim. 9}, 4 (Jan. 2013), 54:1--54:23.

\bibitem{Volkov:GPUs}
{\sc Volkov, V.}
\newblock {\em Understanding Latency Hiding on GPUs}.
\newblock PhD thesis, University of California at Berkeley, 2016.

\bibitem{Wei:DLVM}
{\sc Wei, R., Adve, V., and Schwartz, L.}
\newblock Dlvm: A modern compiler infrastructure for deep learning systems.
\newblock {\em CoRR abs/1711.03016\/} (2017).

\bibitem{Whaley:ATLAS}
{\sc Whaley, R.~C., and Dongarra, J.~J.}
\newblock Automatically tuned linear algebra software.
\newblock In {\em Proceedings of the 1998 ACM/IEEE Conference on
  Supercomputing\/} (Washington, DC, USA, 1998), SC '98, IEEE Computer Society,
  pp.~1--27.

\bibitem{roofline}
{\sc Williams, S., Waterman, A., and Patterson, D.}
\newblock Roofline: An insightful visual performance model for multicore
  architectures.
\newblock {\em Commun. ACM 52}, 4 (Apr. 2009), 65--76.

\bibitem{zaremba2014recurrent}
{\sc Zaremba, W., Sutskever, I., and Vinyals, O.}
\newblock Recurrent neural network regularization.
\newblock {\em arXiv preprint arXiv:1409.2329\/} (2014).

\end{thebibliography}

\end{document}